\documentclass[11pt]{article}
\usepackage{amssymb,amsmath,color,graphicx}
\usepackage{url,times}

\usepackage{amssymb,amsmath,color,times}

\newcommand{\BlackBox}{\rule{1.5ex}{1.5ex}}  % end of proof
\newenvironment{proof}{\par\noindent{\bf Proof\ }}{\hfill\BlackBox\\[2mm]}

\newcommand{\BEAS}{\begin{eqnarray*}}
\newcommand{\EEAS}{\end{eqnarray*}}
\newcommand{\BEA}{\begin{eqnarray}}
\newcommand{\EEA}{\end{eqnarray}}
\newcommand{\BEQ}{\begin{equation}}
\newcommand{\EEQ}{\end{equation}}
\newcommand{\BIT}{\begin{itemize}}
\newcommand{\EIT}{\end{itemize}}
\newcommand{\BNUM}{\begin{enumerate}}
\newcommand{\ENUM}{\end{enumerate}}
\newcommand{\BA}{\begin{array}}
\newcommand{\EA}{\end{array}}
\newcommand{\diag}{\mathop{\rm diag}}
\newcommand{\Diag}{\mathop{\rm Diag}}

\newcommand{\var}{\mathop{ \rm var}}

\newcommand{\tr}{\mathop{ \rm tr}}
\newcommand{\sign}{\mathop{ \rm sign}}
\newcommand{\idm}{I}
\newcommand{\rb}{\mathbb{R}}

\def \lmin{ \lambda_{\min}}
\def \lmax{ \lambda_{\max}}

\def \lmaxQ{ \lmax(Q) }

\newtheorem{lemma}{Lemma}
\newtheorem{theorem}{Theorem}
\newtheorem{proposition}{Proposition}

\newcommand{\mysec}[1]{Section~\ref{sec:#1}}
\newcommand{\eq}[1]{Eq.~(\ref{eq:#1})}

 \def \betac{ R }
 \def \betacuni{ S }

\def \E{{\mathbb E}}
\def \P{{\mathbb P}}

\title{Self-concordant analysis for logistic regression}

\author{ \Large{Francis Bach} \\[.5cm]
INRIA - WILLOW Project-Team \\
Laboratoire d'Informatique de l'Ecole Normale Sup\'erieure \\
(CNRS/ENS/INRIA UMR 8548) \\
23, avenue d'Italie,
75214 Paris, France \\
\texttt{francis.bach@inria.fr}
       }

\begin{document}
\maketitle
\begin{abstract}
Most of the non-asymptotic theoretical work in regression is carried out for the square loss, where estimators can be obtained through closed-form expressions. In this paper, we use and extend  tools from the convex optimization literature, namely self-concordant functions, to provide simple extensions of theoretical results for the square loss to the logistic loss. We apply the extension techniques to logistic regression with regularization by the $\ell_2$-norm and regularization by the $\ell_1$-norm, showing that new results for binary classification through logistic regression can be easily derived from corresponding results for least-squares regression.

\end{abstract}

\section{Introduction}
The theoretical analysis of statistical methods is usually greatly simplified when the estimators have   closed-form expressions. For methods based on the minimization of a certain functional, such as M-estimation methods~\cite{VanDerVaart}, this is true when the function to minimize is quadratic, i.e., in the context of regression, for the square loss.

When such loss is used, asymptotic and non-asymptotic results may be derived with classical tools from probability theory (see, e.g.,~\cite{massart-concentration}). When the function which is minimized in M-estimation is not amenable to closed-form solutions, local approximations are then needed for obtaining and analyzing a solution of the optimization problem. In the asymptotic regime, this has led to interesting developments and extensions of results from the quadratic case, e.g., consistency or asymptotic normality (see, e.g.,~\cite{VanDerVaart}). However, the situation is different when one wishes to derive non-asymptotic results, i.e., results where all constants of the problem are explicit. Indeed, in order to prove results as sharp as for the square loss, much notation and many assumptions have to be introduced regarding second and third derivatives; this makes the derived results much more complicated than the ones for closed-form estimators~\cite{van2008high,gu1990adaptive, bunea2008honest}.

A similar situation occurs in convex optimization, for the study of Newton's method for obtaining solutions of unconstrained optimization problems. It is known to be locally quadratically convergent for convex problems. However, its classical analysis requires cumbersome notations and assumptions regarding second and third-order derivatives~(see, e.g.,~\cite{bertsekas,boyd}). This situation was greatly enhanced with the introduction of the notion of \emph{self-concordant functions}, i.e., functions whose third derivatives are controlled by their second derivatives. With this tool, the analysis is much more transparent~\cite{boyd,self}. While Newton's method is a commonly used algorithm  for logistic regression~(see, e.g.,~\cite{christensen1997log,hosmer2004applied}), leading to iterative least-squares algorithms, we don't focus in the paper on the resolution of the optimization problems, but on the statistical analysis of the associated global minimizers.

In this paper, we aim to borrow tools from convex optimization and self-concordance to analyze the statistical properties of logistic regression. Since the logistic loss is not itself a self-concordant function, we introduce in \mysec{self} a new type of functions with a different control of the third derivatives. For these functions, we prove 
two types of results: first, we provide lower and upper Taylor expansions, i.e., Taylor expansions which are globally upper-bounding or lower-bounding a given function. Second, we prove results on the behavior of Newton's method which are similar to the ones for self-concordant functions. We then apply them in Sections~\ref{sec:log}, \ref{sec:L2} and~\ref{sec:L1} to the one-step  Newton iterate from the population solution of the corresponding problem (i.e., $\ell_2$ or $\ell_1$-regularized logistic regression). This essentially shows that the analysis of logistic regression can be done \emph{non-asymptotically} using the local quadratic approximation of the logistic loss,
\emph{without} complex additional assumptions. Since this approximation corresponds to a weighted least-squares problem, results from least-squares regression can thus be naturally extended.

In order to consider such extensions and make sure that the new results closely match the corresponding ones for least-squares regression, we derive in Appendix~\ref{app:concentration} new Bernstein-like concentration inequalities for quadratic forms of bounded random variables, obtained from general results on U-statistics~\cite{reynaud}.

We first apply in \mysec{L2} the extension technique to regularization by the $\ell_2$-norm, where we consider two settings, a situation  with no assumptions regarding the conditional distribution of the observations, and another one where the model is assumed well-specified and we derive asymptotic expansions of the generalization performance with explicit bounds on remainder terms. In \mysec{L1}, we consider regularization by the $\ell_1$-norm and extend two known recent results for the square loss, one on model consistency~\cite{Zhaoyu,yuanlin,zou,martin} and one on prediction efficiency~\cite{tsyb}. The main contribution of this paper is to make these extensions as simple as possible, by allowing the use of non-asymptotic second-order Taylor expansions.

\paragraph{Notation.}

For $x \in \rb^p$ and $q\geqslant 1$, we  denote by $\| x\|_q$ the $\ell_q$-norm
of $x$, defined as
$\| x\|_q^q = \sum_{i=1}^p |x_i|^q$. We also denote by $\| x\|_\infty =
\max_{i \in \{1,\dots,p\}} | x_i |$ its $\ell_\infty$-norm.
 We   denote by $\lmaxQ$ and $\lambda_{\min}(Q)$ the largest and smallest eigenvalue of a symmetric matrix $Q$.  We use the notation $Q_1 \preccurlyeq Q_2$ (resp. $Q_1 \succcurlyeq Q_2$) for the positive semi-definiteness of the matrix $Q_2-Q_1$ (resp. $Q_1-Q_2$).
 
For   $a \in \rb$, $\sign(a)$ denotes the sign of $a$, defined as $\sign(a)=1$ if $a>0$, $-1$ if $a<0$, and $0$ if $a=0$. For a vector $v \in \rb^p$, $\sign(v) \in \{-1,0,1\}^p$ denotes the   vector of signs of elements of $v$. 
 
 Moreover, given a vector $v \in \rb^p$ and a subset $I$ of $\{1,\dots,p\}$, 
 $|I|$ denotes the cardinal of the set $I$, $v_I$ denotes the vector in $\rb^{|I|}$ of elements of $v$ indexed by $I$. Similarly, for a matrix $A \in \rb^{p \times p}$, $A_{IJ}$  denotes the submatrix of  $A$ composed of elements of $A$ whose rows are in $I$ and columns are in $J$.  
Finally, we let denote $\P$ and $\E$ general probability measures and expectations.

\section{Taylor expansions and Newton's method}
\label{sec:self}

In this section, we consider a generic function $F: \rb^p \to \rb$, which is convex and three times differentiable. We denote by $F'(w) \in \rb^p$ its gradient at $w \in \rb^p$, by $F''(w) \in \rb^{p \times p}$ its Hessian at $w \in \rb^p$. We denote by $\lambda(w) \geqslant 0 $ the smallest eigenvalue of the Hessian $F''(w)$ at $w \in \rb^p$.

If $\lambda(w) >0$, i.e., the Hessian is invertible at $w$, we can define the \emph{Newton step} as $\Delta^N(w) = - F''(w)^{-1} F'(w)$, and the \emph{Newton decrement} $\nu(F,w)$ at $w$, defined through:
$$
\nu(F,w)^2 = F'(w)^\top F''(w)^{-1} F'(w) 
=  \Delta^N(w) ^\top F''(w) \Delta^N(w) .
$$
The \emph{one-step Newton iterate} $w + \Delta^N(w)$ is the minimizer of the second-order Taylor expansion of $F$ at $w$, i.e., of the function $ v \mapsto F(w) + F'(w)(v-w) + \frac{1}{2} ( v-w )^\top F''(w) ( v - w )$. Newton's method consists in successively applying the same iteration until convergence. For more background and details about Newton's method, see, e.g.,~\cite{boyd,bertsekas,bonnans}.

\subsection{Self-concordant functions}

\label{sec:selfconcordant}
We now review some important properties of self-concordant functions~\cite{boyd,self}, i.e., three times differentiable convex functions such that for all $u,v \in \rb^p$, the function $g:t \mapsto F(u+tv)$ satisfies for all $t \in \rb$, $| g'''(t) | \leqslant 2 g''(t)^{3/2}$.

The local behavior of self-concordant functions is well-studied and lower and upper Taylor expansions can be derived (similar to the ones we derive in Proposition~\ref{prop:multivariate}). Moreover, bounds are available for the behavior of Newton's method; given a self-concordant function $F$, if $w \in \rb^p$ is such that $\nu(F,w) \leqslant 1/4$, then $F$ attains its unique global minimum at some $w^\ast \in \rb^p$, and we have the following bound on the error $w-w^\ast$ (see, e.g.,~\cite{self}):
\BEQ
\label{eq:mumu}
(w - w^\ast)^\top {F''(w)} (w - w^\ast) \leqslant 4\nu(F,w)^2.
\EEQ
Moreover, the newton decrement at the one-step Newton iterate from $w \in \rb^p$ can be upper-bounded as follows:
\BEQ
\label{eq:nunu}
\nu(F,w + \Delta^N(w)) \leqslant \nu(F,w)^2,
\EEQ
which allows to prove an upper-bound of the error of the one-step iterate, by application of \eq{mumu} to $w + \Delta^N(w)$.
Note that these bounds are not the sharpest, but are sufficient in our context.
These are commonly used  to show the global convergence of the damped Newton's method~\cite{self} or of Newton's method with backtracking line search~\cite{boyd}, as well as a precise upper bound on the number of iterations to reach a given precision. 

Note that in the context of machine learning and statistics, self-concordant functions have been used for bandit optimization and online learning~\cite{abernethy2008competing}, but for barrier functions related to constrained optimization problems, and not directly for M-estimation.

\subsection{Modifications of self-concordant functions}
\label{sec:newtype}
The logistic function $u\mapsto \log(1+e^{-u})$ is not self-concordant as the third derivative is bounded by a constant times the second derivative (without the power $3/2$). However, similar bounds can be derived with a different control of the third derivatives.  Proposition \ref{prop:multivariate} provides lower and upper Taylor expansions while Proposition \ref{prop:multivariate-bis} considers the behavior of Newton's method. Proofs may be found in Appendix~\ref{app:opt} and follow closely the ones for regular self-concordant functions found in~\cite{self}.

\begin{proposition}[Taylor expansions]
\label{prop:multivariate}
Let $F: \rb^p \mapsto \rb$ be a convex three times differentiable function such that 
 for all $w,v \in \rb^p $, the function  $g(t) = F(w+tv)$ satisfies for all $t \in \rb$,
$|g'''(t)|   
\leqslant \betac  \| v \|_2    \times g''(t)
$, for some $\betac \geqslant 0$.
We then have for all $w,v,z \in \rb^p$:
\BEA
\label{eq:lowerboundF}
  F(w+v) &\!\!\geqslant \!\!&F(w)  + v^\top F'(w)      +  \frac{ v^\top F''(w) v}{\betac^2 \| v \|_2^2   }
(e^{-\betac \| v \|_2  } + \betac \| v \|_2  - 1),
\\
\label{eq:upperboundF}
  F(w+v) &\!\!\leqslant\!\!& F(w)  + v^\top F'(w)      +  \frac{ v^\top F''(w) v}{\betac^2 \| v \|_2^2  }
(e^{\betac \| v \|_2  } - \betac \| v \|_2   - 1),
\EEA
\BEQ
  \label{eq:boundgradientF}
  \frac{ z ^\top [F'(w+v) \!-\! F'(w)\! -\! F''(w) v  ] }{ [  z^\top F''(w) z]^{1/2} } \leqslant
  [  v^\top F''(w) v]^{1/2} \frac{ e^{  \betac \| v\|_2} \!-\! 1 \!- \!\betac \| v\|_2}{  \betac \| v\|_2 },
  \EEQ
   \BEQ
  \label{eq:boundhessianF}
e^{-\betac \| v \|_2 } F''(w)  \preccurlyeq   F''(w+v) \preccurlyeq  
e^{\betac \| v \|_2 } F''(w)  .
\EEQ
\end{proposition}
Inequalities in \eq{lowerboundF} and \eq{upperboundF} provide upper and lower second-order Taylor expansions of $F$, while \eq{boundgradientF} provides a first-order Taylor expansion of $F'$ and \eq{boundhessianF} can be considered as an upper and lower zero-order Taylor expansion of $F''$. Note the difference here between Eqs.~(\ref{eq:lowerboundF}-\ref{eq:upperboundF}) and regular third-order Taylor expansions of $F$: the remainder term in the Taylor expansion, i.e., 
$F(w+v) - F(w)  - v^\top F'(w)      -  \frac{1}{2} v^\top F''(w) v $ is upper-bounded by 
$ \frac{ v^\top F''(w) v}{\betac^2 \| v \|_2^2  }
(e^{\betac \| v \|_2  } -  \frac{1}{2}\betac^2 \| v \|_2^2- \betac \| v \|_2   - 1)
$; for $\| v\|_2$ small, we obtain a term proportional to $\|v\|_2^3$ (like a regular local Taylor expansion), but the bound remains valid for all $v$ and does not grow as fast as a third-order polynomial. Moreover, a regular Taylor expansion with a uniformly bounded third-order derivative would lead to a bound proportional to
$ \|v\|_2^3$, which does not take into account the local curvature of $F$ at $w$.
Taking into account this local curvature is key  to obtaining sharp and simple bounds on the behavior of Newton's method (see proof in Appendix~\ref{app:opt}):
\begin{proposition}[Behavior of Newton's method]
\label{prop:multivariate-bis}
Let $F: \rb^p \mapsto \rb$ be a convex three times differentiable function such that 
 for all $w,v \in \rb^p $, the function  $g(t) = F(w+tv)$ satisfies for all $t \in \rb$,
$|g'''(t)|   
\leqslant \betac  \| v \|_2    \times g''(t)
$, for some $\betac \geqslant 0$.
Let $\lambda(w) >0$ be the lowest eigenvalue of $F''(w)$ for some $w \in \rb^p$. If $\nu(F,w) \leqslant \frac{\lambda(w)^{1/2}} {2 \betac      } $, then $F$ has a unique global minimizer $w^\ast \in \rb^p$ and we have:
 \BEA
\label{eq:nu2}
\big(w    - w^\ast \big)^\top {F''(w)} \big(w   - w^\ast\big)  & \leqslant &
16 \nu(F,w)^2, \\
\label{eq:nunew}
\frac{R \nu(F,w + \Delta^N(w)) }{\lambda(w + \Delta^N(w) )^{1/2}}
& \leqslant &  \bigg( \frac{R \nu(F,w)}{ \lambda(w )^{1/2}}
\bigg)^2,
%\\
%\label{eq:nu2BB}
%F(w) - F(w^\ast) & \leqslant &     \nu(F,w)^2,
\\
\label{eq:nu4}
\big(w + \Delta^N(w)  - w^\ast\big)^\top {F''(w)} \big(w +  \Delta^N(w) - w^\ast\big)  & \leqslant & 
\frac {16 \betac^2 }  {\lambda(w)} \nu(F,w)^4.
% \\
%\label{eq:nu4BB}
% F(w + \Delta^N(w)  ) - F(w^\ast) & \leqslant &  \frac { \betac^2 }  {\lambda(w)} \nu(F,w)^4.
 \EEA
\end{proposition}
\eq{nu2} extends \eq{mumu} while \eq{nunew} extends \eq{nunu}. Note that the notion and the results are not invariant by affine transform (contrary to self-concordant functions) and that we still need a (non-uniformly) lower-bounded Hessian. The last two propositions constitute the main technical contribution of this paper. We now apply these to logistic regression and its regularized versions.

\section{Application to logistic regression}
\label{sec:log}
We consider $n$ pairs of observations $(x_i,y_i)$ in $\rb^p \times \{-1,1\}$ and the following objective function for logistic regression:
\BEQ
\label{eq:primal}
\hat{J}_0(w)  = \frac{1}{n} \sum_{i=1}^n  \log \left(
 1 +\exp(-y_i w^\top x_i) \right)  =   \frac{1}{n} \sum_{i=1}^n \left\{  \ell(w^\top x_i) - \frac{y_i}{2} w^\top x_i \right\}  ,
\EEQ
where $  \ell: u \mapsto \log  ({e^{-u/2} + e^{u/2}})  $ is an even convex function. A short calculation leads to  
$\ell'(u) = - 1/2 +  \sigma(u)$,
$\ell''(u) =   \sigma(u)[ 1 - \sigma(u) ]$, 
$\ell'''(u) =   \sigma(u)[ 1 - \sigma(u) ][ 1 - 2\sigma(u) ]$, where
$\sigma(u) = (1+e^{-u})^{-1}$ is the sigmoid function. Note that we have for all $u\in \rb$, $| \ell'''(u)| \leqslant \ell''(u)$. The cost function $\hat{J}_0$
defined in \eq{primal} is proportional to the negative conditional log-likelihood of the data under the conditional model
$\P(y_i=\varepsilon_i|x_i) = \sigma(\varepsilon_i w^\top x_i)$.

If $R = \max_{ i \in \{1,\dots,n \}} \| x_i\|_2$ denotes the maximum $\ell_2$-norm of all input data points, then the cost function $\hat{J}_0$ defined in \eq{primal} satisfies the assumptions of Proposition~\ref{prop:multivariate-bis}. Indeed, we have, with the notations of Proposition~\ref{prop:multivariate-bis},
\BEAS
| g'''(t) |  & = &  \bigg| 
\frac{1}{n} \sum_{i=1}^n \ell'''[(w+tv)^\top x_i]  (x_i^\top v)^3 
\bigg| \\
& \leqslant & 
\frac{1}{n} \sum_{i=1}^n \ell''[(w+tv)^\top x_i]   (x_i^\top v)^2 \|v\|_2 \|x_i\|_2
 \leqslant R \|v \|_2 \times  g''(t).
\EEAS
 Throughout this paper, we will consider a certain vector $w \in \rb^p$ (usually defined through the population functionals) and consider the one-step Newton iterate from this~$w$. Results from \mysec{newtype} will allow to show that this approximates the global minimum of $\hat{J}_0$ or  a regularized version thereof.

Throughout this paper, we consider a \emph{fixed design} setting (i.e., $x_1,\dots,x_n$ are consider deterministic) and we make the following assumptions:
\BIT

\item[\textbf{(A1)}] \emph{Independent outputs}:
The outputs $y_i \in \{-1,1\}$, $i=1,\dots,n$ are independent (but not identically distributed).
 
%\item[\textbf{(A2)}] \emph{Bounded output distributions}: $\max_{i \in \{1,\dots,n\}} | \E  y_i| < 1/2$.

\item[\textbf{(A2)}] \emph{Bounded inputs}:  
$\max_{i \in \{1,\dots,n\}} \| x_i\|_2
\leqslant R$.
 \EIT

We define the model as \emph{well-specified} if there exists $w_0 \in \rb^p$ such that for all $i=1,\dots,n$, $\P(y_i \!=\! \varepsilon_i) = \sigma(  \varepsilon_i w_0^\top x_i)$, which is equivalent to $ \E (y_i /2) = \ell'(w_0^\top x_i)$, and  implies  $ \var (y_i/2) = \ell''(w_0^\top x_i)$. However, we do not always make such assumptions in the paper.

We use the matrix notation $X = [ x_1,\dots,x_n ]^\top \in \rb^{ n \times p}$ for the design matrix and $\varepsilon_i = y_i/2 - \E  (y_i /2)$, for $i=1,\dots,n$, which formally corresponds to the additive noise in least-squares regression. 
 We also use the notation
 % $P_\lambda = J_\lambda''(w_\lambda) =  \frac{1}{n} X^\top \Diag(\ell''(w_\lambda^\top x_i) ) X \in \rb^{p \times p} $,
 $Q = \frac{1}{n} X^\top \Diag( \var(y_i/2) ) X \in \rb^{p \times p} $
  and $q = \frac{1}{n}X^\top \varepsilon \in \rb^{p } $. By assumption, we have $\E (qq^\top) = \frac{1}{n} Q$.

We denote by $J_0$ the expectation of $\hat{J}_0$, i.e.:
$$
 {J}_0(w) = \E   \big[\hat{J}_0(w)  \big] = \frac{1}{n} \sum_{i=1}^n 
  \left\{  \ell(w^\top x_i) -  \E(y_i/2)  w^\top x_i \right\}.
  $$
  Note that with our notation, $\hat{J}_0(w) = J_0(w) - q^\top w$.
In this paper we consider $J_0(\hat{w})$ as the generalization performance of a certain estimator $\hat{w}$. This corresponds to the average Kullback-Leibler divergence to the best model when the model is well-specified, and is common for the study of logistic regression and more generally generalized linear models~\cite{mccullagh1989generalized, efron2004estimation}. Measuring the classification performance through the 0--1 loss~\cite{bartlett2006convexity} is out of the scope of this paper.

 The function $J_0$ is bounded from below, therefore it has a bounded infimum $\inf_{w \in \rb^p} J_0(w) \geqslant 0 $. This infimum might or might not be attained at a finite $w_0 \in \rb^p$; when the model is well-specified, it is always attained (but this is not a necessary condition), and, unless the design matrix $X$ has rank $p$, is not unique.
 
 The difference between the analysis through self-concordance and the classical asymptotic analysis is best seen when the model is well-specified, and exactly mimics the difference between self-concordant analysis of Newton's method and its classical analysis. The usual analysis of logistic regression requires that the logistic function $u \mapsto \log(1+e^{-u})$ is strongly convex (i.e., with a strictly positive lower-bound on the second derivative), which is true only on a compact subset of $\rb$. Thus, non-asymptotic results such as the ones from \cite{bunea2008honest,van2008high} requires an upper bound $M$ on $|w_0^\top x_i|$, where $w_0$ is the generating loading vector; then, the second derivative of the logistic loss is lower bounded by $(1 + e^M)^{-1}$, and this lower bound may be very small when $M$ gets large. Our analysis does not require such a bound because of the fine control of the third derivative.

\section{Regularization by the $\ell_2$-norm}
\label{sec:L2}
We denote by $\hat{J}_\lambda(w) = \hat{J}_0(w) + \frac{\lambda}{2} \| w\|_2^2$ the empirical  $\ell_2$-regularized functional. For $\lambda>0$, the function
$\hat{J}_\lambda$ is strongly convex and we denote by 
$\hat{w}_\lambda$ the unique global minimizer of  $\hat{J}_\lambda$. In this section, our goal is to find upper and lower bounds on the generalization performance $J_0(\hat{w}_\lambda)$, under minimal assumptions (\mysec{minimal}) or when the model is well-specified~(\mysec{wellspecified}).

\subsection{Reproducing kernel Hilbert spaces and splines}
\label{sec:rkhs}
In this paper we focus explicitly on \emph{linear} logistic regression, i.e., on a generalized linear model that allows linear dependency between $x_i$ and the distribution of~$y_i$. Although apparently limiting, in the context of regularization by the $\ell_2$-norm, this setting contains \emph{non-parametric} and \emph{non-linear} methods based on splines or reproducing kernel Hilbert spaces (RKHS)~\cite{wahba}. Indeed, because of the representer theorem~\cite{representer},  minimizing the cost function
$$
  \frac{1}{n} \sum_{i=1}^n \left\{  \ell[ f(x_i) ] - \frac{y_i}{2} f(x_i) \right\} 
  + \frac{\lambda}{2} \| f\|_\mathcal{F}^2,
$$
   with respect to the function $f$ in the RKHS $\mathcal{F}$ (with norm $\| \cdot \|_\mathcal{F}$ and kernel $k$), is equivalent to minimizing the   cost function
   \BEQ
   \label{eq:rkhs}
     \frac{1}{n} \sum_{i=1}^n \left\{  \ell[ (T \beta)_i ] - \frac{y_i}{2} (T \beta)_i\right\} 
  + \frac{\lambda}{2} \| \beta \|_2^2,
\EEQ
with respect to $\beta \in \rb^p$,
where $T \in \rb^{n \times p }$ is a square root of the kernel matrix $K \in \rb^{n \times n}$ defined as $K_{ij} = k(x_i,x_j)$, i.e., such that $K = TT^\top$. The unique solution of the original problem $f$ is then obtained as
$f(x) = \sum_{i=1}^n \alpha_i k(x,x_i)$, where $\alpha$ is any vector satisfying $ TT^\top \alpha = T \beta$ (which can be obtained by matrix pseudo-inversion~\cite{golub83matrix}). Similar developments can be carried out for smoothing splines~(see, e.g.,~\cite{wahba,gu2002smoothing}). By identifying the matrix $T$ with the data matrix $X$, the optimization problem in \eq{rkhs} is identical to minimizing $\hat{J}_0(w) + \frac{\lambda}{2}\| w\|_2^2$,  and thus our results apply to estimation in RKHSs.

   \subsection{Minimal assumptions (misspecified model)}
\label{sec:minimal}
In this section, we do not assume that the model is well-specified. We obtain the following theorem (see proof in Appendix~\ref{app:mis}), which only assumes boundedness of the covariates and independence of the outputs:
\begin{theorem}[Misspecified model]
\label{theo:mis}
Assume \textbf{(A1)}, \textbf{(A2)} and $\lambda =  { 19 R^2} \sqrt{\frac{\log(8/\delta)}{n}}$, with $\delta  \in (0,1)$. Then, with probability at least $1-\delta$, for all $w_0 \in \rb^p$,
\BEQ
\label{eq:mis}
J_0(\hat{w}_\lambda) \leqslant J_0(w_0)  +    \big( 10 + 100 R^2 \|w_0\|_2^2 \big) \sqrt{\frac{\log(8/\delta)}{n}}.
\EEQ
\end{theorem}

In particular, if the global minimum of $J_0$ is attained at $w_0$ (which is not an assumption of Theorem~\ref{theo:mis}), we obtain an oracle inequality as $J_0(w_0) = \inf_{w \in \rb^p} J_0(w)$. The lack of additional assumptions unsurprisingly gives rise to a slow rate of~$n^{-1/2}$.

This is to be compared with~\cite{sridharan2008fast}, which uses  different proof techniques but obtains similar results for all convex Lipschitz-continuous losses (and not only for the logistic loss). However, the techniques presented in this paper allow the derivation of much more precise statements in terms of bias and variance (and with better rates), that involves some knowledge of the problem. We do not pursue detailed results here, but focus in the next section on well-specified models, where results have a simpler form.
 
This highlights two opposite strategies for the theoretical analysis of regularized problems: the first one, followed by~\cite{sridharan2008fast,steinwart2006new}, is mostly loss-independent and relies on advanced tools from empirical process theory, namely uniform concentration inequalities. Results are widely applicable and make very few assumptions. However, they tend to give performance guarantees which are far below the observed performances of such methods in applications. The second strategy, which we follow in this paper, is to restrict the loss class (to linear or logistic) and derive the limiting convergence rate, which does depend on unknown constants (typically the best linear classifier itself). Once the limit is obtained, we believe it gives a better interpretation of the performance of these methods, and if one really wishes to make no assumption, taking upper bounds on these quantities, we may get back results obtained with the generic strategy, which is exactly what Theorem~\ref{theo:mis} is achieving.

Thus, a detailed analysis of the convergence rate, as done in Theorem~\ref{theo:perf} in the next section, serves two purposes: first, it gives a sharp result that depends on unknown constants; second the constants can be maximized out and more general results may be obtained, with fewer assumptions but worse convergence rates.

 \subsection{Well-specified models}
\label{sec:wellspecified}
 We now assume that the model is well-specified, i.e., that the probability that $y_i=1$ is a sigmoid function of a linear function of $x_i$, which is equivalent to:
 \BIT

\item[\textbf{(A3)}] \emph{Well-specified model}: There exists $w_0 \in \rb^p$ such that $ \E ( y_i /2) = \ell'(w_0^\top x_i)$.
\EIT

Theorem~\ref{theo:perf} will give upper and lower bounds on the expected risk of the $\ell_2$-regularized estimator $\hat{w}_\lambda$, i.e., $J_0(\hat{w}_\lambda)$. We use the following definitions for the two degrees of freedom and biases, which are usual in the context of ridge regression and spline smoothing~(see, e.g.,~\cite{wahba,gu2002smoothing,minikernel}):
$$\begin{array}{llcl}
\mbox{ degrees of freedom (1)} : & d_1 & =  &\tr Q ( Q+ \lambda \idm)^{-1} ,  \\
\mbox{ degrees of freedom (2)}  : & d_2  & = &  \tr Q^2 ( Q+ \lambda \idm)^{-2}  ,    \\
\mbox{ bias (1)} :  & b_1 &  = &  \lambda^2 w_0^\top ( Q+ \lambda \idm)^{-1} w_0 ,   \\
\mbox{ bias (2)} :  & b_2 & =  & \lambda^2 w_0^\top Q ( Q+ \lambda \idm)^{-2} w_0.
\end{array}
$$
Note that we always have the inequalities  $d_2 \leqslant d_1 \leqslant 
\min \{R^2/ \lambda,n\}$ and $b_2 \leqslant b_1 \leqslant \min \{\lambda \|w_0\|_2^2, \lambda^2 w_0^\top Q^{-1} w_0 \}$, and that these quantities depend on $\lambda$. In the context of RKHSs outlined in \mysec{rkhs}, we have $d_1 = \tr K ( K + n \lambda \Diag(\sigma_i^2))^{-1}$, a quantity which is also usually referred to as the \emph{degrees of freedom}~\cite{hastie_GAM}. In the context of the analysis of $\ell_2$-regularized methods,  the two degrees of freedom are necessary, as outlined in Theorems~\ref{theo:perf} and \ref{theo:smooth}, and in \cite{minikernel}.

Moreover, we denote by $\kappa > 0 $ the following quantity 
\BEQ
\label{eq:kappa2}
\kappa = \frac{R}{\lambda^{1/2}} \bigg( \frac{d_1}{n} + b_1 \bigg)
  \bigg( \frac{d_2}{n} + b_2 \bigg)^{-1/2} \!\! .
\EEQ
Such quantity is an extension of the one used by~\cite{Harchaoui:Bach:Moulines:2008} in the context of kernel Fisher discriminant analysis used as a test for homogeneity.  In order to obtain asymptotic equivalents, we require $\kappa$ to be small, which, as shown later in this section, occurs in many interesting cases when $n$ is large enough.

In this section, we will apply
results from \mysec{self} to the functions $\hat{J}_\lambda$ and $J_0$. Essentially, we will consider local quadratic approximations
of these functions around the generating loading vector $w_0$, leading to replacing the true estimator $\hat{w}_\lambda$ by the one-step Newton iterate from $w_0$. This is only possible if the Newton decrement
  $\nu(\hat{J}_\lambda,w_0)$ is small enough, which  leads to additional constraints (in particular the upper-bound on $\kappa$).

\begin{theorem}[Asymptotic generalization performance]
\label{theo:perf}
Assume \textbf{(A1)}, \textbf{(A2)} and  \textbf{(A3)}. Assume moreover $\kappa \leqslant 1/16$, where $\kappa$ is defined in \eq{kappa2}. If $v \in [0,1/4]$ satisfies $v^3 ( d_2 + nb_2)^{1/2} \leqslant 12$, then, with probability at least $1-\exp( - v^2 ( d_2 + nb_2) )$:
\BEQ
\label{eq:perf}
 \bigg | J_0(\hat{w}_\lambda) -  J_0(w_0) - \frac{1}{2} \bigg(b_2 + \frac{d_2}{n} \bigg) 
 \bigg| \leqslant \bigg( b_2 + \frac{d_2}{n} \bigg)  ( 69 v + 2560 \kappa).
\EEQ
\end{theorem}

\paragraph{Relationship to previous work.} When the dimension $p$ of $w_0$ is bounded, then under the regular asymptotic regime ($n$ tends to $+\infty$), $J_0(\hat{w}_\lambda)$ has the following expansion $ J_0(w_0) + \frac{1}{2} \big(b_2 + \frac{d_2}{n} \big)$, a result which has been obtained by several authors in several settings~\cite{shibata, bozdogan1987model}. In this asymptotic regime, the optimal~$\lambda$ is known to be of order $O(n^{-1})$~\cite{percy}. The main contribution of our analysis is to allow a non asymptotic analysis with explicit constants. Moreover, note that for the square loss, the bound in \eq{perf} holds with $\kappa=0$, which can be linked to the fact that our self-concordant analysis from
Propositions~\ref{prop:multivariate} and \ref{prop:multivariate-bis} is applicable with $R = 0$  for the square loss. Note that the constants in the previous theorem could probably be improved.

\paragraph{Conditions for asymptotic equivalence.}
In order to have the remainder term in \eq{perf} negligible with high probability compared to the lowest order term in the expansion of $J_0(\hat{w}_{\lambda})$, we need to have $d_2 + n b_2$  large and $\kappa$ small (so that $v$ can be taken taking small while $ v^2 ( d_2 + nb_2)$ is large, and hence we have a result with high-probability).  The assumption that $d_2 + n b_2$ grows unbounded when $n$ tends to infinity is a classical assumption in the study of smoothing splines and RKHSs~\cite{Cra_Wah:1979,KCLi:1987}, and simply states that the convergence rate of the excess risk $J_0(\hat{w}_\lambda)-J_0(w_0)$, i.e., $b_2 + d_2/n$, is slower than for parametric estimation, i.e., slower than $n^{-1}$.

\paragraph{Study of parameter $\kappa$.} First, 
we always have $\kappa \geqslant  \frac{R}{\lambda^{1/2}} \big( \frac{d_1}{n} + b_1 \big)
^{1/2}$; thus an upper bound on $\kappa$ implies an upperbound on $ \frac{d_1}{n} + b_1 $ which is needed in the proof of Theorem~\ref{theo:perf} to show that the Newton decrement is small enough.
Moreover, $\kappa$ is bounded by the sum of $\kappa_{\rm bias} = 
\frac{R}{\lambda^{1/2}}   b_1 
    b_2  ^{-1/2}$ and 
  $\kappa_{\rm var} = 
\frac{R}{\lambda^{1/2}} \big( \frac{d_1}{n}   \big)
  \big( \frac{d_2}{n}   \big)^{-1/2}$.
Under simple assumptions on the eigenvalues of $Q$ or equivalently of $\Diag(\sigma_i) K \Diag(\sigma_i)$, one can show that $\kappa_{\rm var}$ is small. For example, if $d$ of these eigenvalues are equal to one and the remaining ones are zero, then, 
$
\kappa_{\rm var} = \frac{R d^{1/2}}{\lambda^{1/2} n^{1/2}}  $. And thus we simply need $\lambda$ asymptotically greater than $R^2 d /n $. For additional conditions for $\kappa_{\rm var}$, see~\cite{minikernel,Harchaoui:Bach:Moulines:2008}. A simple condition for $\kappa_{\rm bias}$ can be obtained if $w_0^\top Q^{-1} w_0$ is assumed bounded (in the context of RKHSs this is a stricter condition that the generating function is inside the RKHS, and is used by~\cite{grouplasso} in the context of sparsity-inducing norms). In this case, the bias terms are negligible compared to the variance term as soon as $\lambda$ is asymptotically greater than $n^{-1/2}$.

\paragraph{Variance term.} Note that the diagonal matrix $\Diag(\sigma_i^2)$ is upperbounded by $\frac{1}{4}\idm$, i.e., $\Diag(\sigma_i^2) \preccurlyeq \frac{1}{4} \idm$, so that the degrees of freedom for logistic regression are always less than the corresponding ones for least-squares regression (for $\lambda$ multiplied by 4). Indeed, the pairs $(x_i,y_i)$ for which the conditional distribution is close to deterministic are such that $\sigma_i^2$ is close to zero. And thus it should reduce the variance of the estimator, as little noise is associated with these points, and the effect of this reduction is exactly measured by the reduction in the degrees of freedom.

Moreover, the rate of convergence $d_2/n$ of the variance term has been studied by many authors~(see, e.g., \cite{wahba,gu2002smoothing,Harchaoui:Bach:Moulines:2008}) and depends on the decay of the eigenvalues of $Q$  (the faster the decay, the smaller $d_2$). The degrees of freedom usually grows with $n$, but in many cases is slower than $n^{1/2}$, leading to faster rates in \eq{perf}.

\subsection{Smoothing parameter selection}

In this section,  we obtain a criterion similar to Mallow's $C_L$~\cite{Mal:1973} to estimate the generalization error and select in a data-driven way the regularization parameter~$\lambda$ (referred to as the smoothing parameter when dealing with splines or RKHSs). The following theorem shows that with a data-dependent   criterion, we may obtain a good estimate of the generalization performance, up to a constant term $q^\top w_0$ independent of $\lambda$ (see proof in Appendix~\ref{app:smooth}):
\begin{theorem}[Data-driven estimation of generalization performance]
\label{theo:smooth}
Assume \textbf{(A1)}, \textbf{(A2)} and  \textbf{(A3)}. Let $\hat{Q}_\lambda = \frac{1}{n} \sum_{i=1}^n \ell''(\hat{w}_\lambda^\top x_i) x_i x_i^\top$
and $q = \frac{1}{n} \sum_{i=1}^n ( y_i/2 - \E(y_i/2) ) x_i$.
 Assume moreover $\kappa \leqslant 1/16$, where $\kappa$ is defined in \eq{kappa2}. If $v \in [0,1/4]$ satisfies $v^3 ( d_2 + nb_2)^{1/2} \leqslant 12$, then, with probability at least $1-\exp( - v^2 ( d_2 + nb_2) )$:
$$
\bigg|  J_0(\hat{w}_\lambda) - \hat{J}_0(\hat{w}_\lambda) - \frac{1}{n} \tr \hat{Q}_\lambda ( \hat{Q}_\lambda  + \lambda \idm)^{-1} - q^\top w_0 \bigg| \leqslant   \bigg( b_2 + \frac{d_2}{n} \bigg)    ( 69 v + 2560 \kappa).
$$
\end{theorem}
The previous theorem, which is essentially a non-asymptotic version of results in~\cite{shibata, bozdogan1987model} can be further extended to obtain oracle inequalities when minimizing the data-driven criterion 
$\hat{J}_0(\hat{w}_\lambda) + \frac{1}{n} \tr \hat{Q}_\lambda ( \hat{Q}_\lambda  + \lambda \idm)^{-1}$, similar to results obtained in~\cite{KCLi:1987,minikernel} for the square loss. Note that contrary to   least-squares regression   with Gaussian noise, there is no need to estimate the unknown noise variance (of course only when the logistic model is actually well-specified); however, the matrix $Q$ used to define the degrees of freedom does depend on $w_0$ and thus requires that $\hat{Q}_\lambda$ is used as an estimate. Finally, criteria based on generalized cross-validation~\cite{o1986automatic,gu1990adaptive} could be studied with similar tools.

\section{Regularization by the $\ell_1$-norm}
\label{sec:L1}
In this section, we consider an estimator $\hat{w}_\lambda$ obtained as a minimizer of the $\ell_1$-regularized empirical risk, i.e., 
$\hat{J}_0(w) + \lambda \| w\|_1$. It is well-known that the estimator has some zero components~\cite{lasso}. In this section, we extend some of the recent results~\cite{Zhaoyu,yuanlin,zou,martin,tsyb,zhangL1} for the square loss (i.e., the Lasso) to the logistic loss. We assume throughout this section that the model is well-specified, that is, that the observations $y_i$, $i=1,\dots,n$, are generated according to the logistic model $\P(y_i = \varepsilon_i) = \sigma(\varepsilon_i w_0^\top x_i)$.

 We denote by $K = \{ j \in \{1,\dots,p\}, (w_0)_j \neq 0 \}$ the set of non-zero components of $w_0$ and $s = \sign ( w_0) \in \{-1,0,1\}^p$ the vector of signs of $w_0$. On top of Assumptions  \textbf{(A1)}, \textbf{(A2)} and  \textbf{(A3)}, we will make the following assumption regarding normalization for each covariate (which can always be imposed by renormalization), i.e.,
\BIT

\item[\textbf{(A4)}] \emph{Normalized covariates}: for all $j=1,\dots,p$, 
$\frac{1}{n}\sum_{i=1}^n [(x_i)_j]^2 \leqslant 1$.

\EIT

In this section, we consider two different results, one on model consistency (\mysec{cons}) and one on efficiency (\mysec{eff}). As for the square loss, they will both depend on additional assumptions regarding the square $p \times p$ matrix $Q = \frac{1}{n} \sum_{i=1}^n \ell''(w_0^\top x_i) x_i x_i^\top$.
This matrix is a weighted Gram matrix, which corresponds to the unweighted one for the square loss. As already shown in~\cite{bunea2008honest,van2008high}, usual assumptions for the Gram matrix for the square loss are extended, for the
logistic loss setting using the weighted Gram matrix $Q$. In this paper, we consider two types of results based on specific assumptions on $Q$, but other ones could be considered as well (such as~\cite{juditsky2008verifiable}). The main contribution of using self-concordant analysis is to allow simple extensions from the square loss with short proofs and sharper bounds, in particular by avoiding an exponential constant in the maximal value of $|w_0^\top x_i|$, $i=1,\dots,n$.

\subsection{Model consistency condition}
\label{sec:cons}

The following theorem provides a sufficient condition for model consistency. It is based on the \emph{consistency condition}
$ \| Q_{K^c K}Q_{KK}^{-1} s_K \|_\infty <1$, which is exactly the same as the one for the square loss~\cite{martin,Zhaoyu,zou} (see proof in Appendix~\ref{app:consistency}):
\begin{theorem}[Model consistency for $\ell_1$-regularization]
\label{theo:consistency}
Assume \textbf{(A1)}, \textbf{(A2)}, \textbf{(A3)} and \textbf{(A4)}.
 Assume that there exists $\eta,\rho, \mu >0$ such that
\BEQ
\label{eq:cond}
 \| Q_{K^c K}Q_{KK}^{-1} s_K \|_\infty \leqslant 1 - \eta,
 \EEQ
 $\lmin(Q_{KK} ) \geqslant \rho  $ and $\min_{j \in K } |(w_0)_j|  \geqslant \mu$. Assume $ \lambda \leqslant 
\min \left\{
 \frac{\rho \mu}{ 4 |K|^{1/2} } ,
 \frac{ \eta \rho^{3/2} }{64 R |K|}
 \right\}$.
Then the probability that the vector of signs of $\hat{w}_\lambda$ is different from $s
=\sign(w_0)$ is upperbounded by
\BEQ
\label{eq:bound}
2 p \exp \bigg( - \frac{ n \lambda^2 \eta^2}{16} \bigg)
+ 2|K| \exp \bigg(
-\frac{n\rho^2 \mu^2 }{16|K|}
\bigg) +  2 |K| \exp \bigg(
-\frac{\lambda n\rho^{3/2} \eta  }{64 R |K|}
\bigg) .
\EEQ
\end{theorem}
 
 \paragraph{Comparison with square loss.} For the square loss, the previous theorem simplifies~\cite{martin,Zhaoyu}: with our notations,  the constraint $\lambda \leqslant 
 \frac{ \eta \rho^{3/2} }{64 R |K|}$ and the last term in \eq{bound}, which are the only ones depending on $R$,  can be removed (indeed, the square loss allows the application of our adapted self-concordant analysis with the constant $\betac=0$). On the one hand, the favorable scaling between $p$ and $n$, i.e., $\log p =O(n)$ for a certain well-chosen $\lambda$, is preserved (since   the logarithm of the added term is proportional to $-\lambda n$). However, on the other hand, the terms in $R$ may be large as $R$ is the radius of the entire data (i.e., with all $p$ covariates). Bounds with the radius of the data on only the relevant features in $K$ could be derived as well (see details in the proof in Appendix~\ref{app:consistency}).

\paragraph{Necessary condition.} In the case of the square loss, 
 a weak form of \eq{cond}, i.e., $ \| Q_{K^c K}Q_{KK}^{-1} s_K \|_\infty \leqslant 1$ turns out to be necessary and sufficient  for asymptotic correct model selection~\cite{zou}. While the weak form is clearly necessary for model consistency, and the strict form sufficient (as proved in Theorem~\ref{theo:consistency}), we are currently investigating whether the weak condition is also sufficient for the logistic loss.

\subsection{Efficiency}
\label{sec:eff}

Another type of result has been derived, based on different proof techniques~\cite{tsyb} and aimed at efficiency (i.e., predictive performance). Here again, we can extend the result in a very simple way. We assume, given $ K $ the set of non-zero components of $w_0$:
\BIT

\item[\textbf{(A5)}] \emph{Restricted eigenvalue condition}:  
$$
\rho  =  
\min_{\|\Delta_{K^c}\|_1 \leqslant 3 \| \Delta_{K}\|_1 } \frac{ ( \Delta^\top Q \Delta)^{1/2}}{ \| \Delta_{K} \|_2} >  0 .
$$
 \EIT

Note that the assumption made in~\cite{tsyb} is slightly stronger but only depends on the cardinality of $K$ (by minimizing with respect to all sets of indices with cardinality equal to the one of $K$). The following theorem provides an estimate of the estimation error as well as an oracle inequality for the generalization performance (see proof in Appendix~\ref{app:efficiency}):
\begin{theorem}[Efficiency for $\ell_1$-regularization]
\label{theo:efficiency}
Assume \textbf{(A1)}, \textbf{(A2)}, \textbf{(A3)}, \textbf{(A4)}, and   \textbf{(A5)}. 
For all $\lambda \leqslant \frac{\rho^2}{48 R |K| }$, with probability at least
$1 - 2p e^{- \lambda n^2 / 5 }$, we have:
\BEAS
 \| \hat{w}_\lambda - w_0 \|_1 &\leqslant &12 \lambda |K| \rho^{-2}, \\
   {J}_0(\hat{w}_\lambda) -   {J}_0(w_0) &\leqslant&12 \lambda^2 |K| \rho^{-2}.
\EEAS
\end{theorem}
We obtain a result  which directly mimics the one obtained in~\cite{tsyb} for the square loss with the exception of the added bound on $\lambda$. In particular, if we take $\lambda =    \sqrt{ \frac{ 10 \log(p)}{n}}$, we get with probability at least
$1 - 2/p$, an upper bound on the generalization performance $ {J}_0(\hat{w}_\lambda)    \leqslant  {J}_0(w_0)   + 120  \frac{\log p}{n} |K| \rho^{-2}$. Again, the proof of this result  is a direct extension of the corresponding one for the square loss, with few additional assumptions owing to the proper self-concordant analysis.

\section{Conclusion}
We have provided an extension of self-concordant functions that allows the simple extensions of theoretical results for the square loss to the logistic loss. We have applied the extension techniques to regularization by the $\ell_2$-norm and regularization by the $\ell_1$-norm, showing that new results for logistic regression can be easily derived from corresponding results for least-squares regression, without added complex assumptions.

The present work could be extended in several interesting ways to different settings. First, for logistic regression, other extensions of theoretical results from least-squares regression could be carried out: for example, the analysis of sequential experimental design for logistic regression leads to many assumptions that could be relaxed (see, e.g.,~\cite{chaudhuri1993nonlinear}). Also, other regularization frameworks based on sparsity-inducing norms could be applied to logistic regression with similar guarantees than for least-squares regression, such as group Lasso for grouped variables~\cite{grouped} or non-parametric problems~\cite{grouplasso}, or resampling-based procedures~\cite{bolasso,meinshausen2008stability} that allow to get rid of sufficient consistency conditions.

Second, the techniques developed in this paper could be extended to other M-estimation problems: indeed, other generalized linear models beyond logistic regression could be considered where higher-order derivatives can be expressed through cumulants~\cite{mccullagh1989generalized}. Moreover, similar developments could be made for density estimation for the exponential family, which would in particular lead to interesting developments for Gaussian models in high dimensions, where $\ell_1$-regularization has proved useful~\cite{meinshausen2006high,banerjee}.
Finally, other losses for binary or multi-class classification are of clear interest~\cite{bartlett2006convexity}, potentially with different controls of the third derivatives.

\appendix

\section{Proofs of optimization results}
\label{app:opt}
We follow the proof techniques of~\cite{self}, by simply changing the control of the third order derivative.
We denote by $F'''(w)$ the third-order derivative of $F$, which is itself a function from $\rb^p \times \rb^p \times \rb^p$ to $\rb$. The
assumptions made in Propositions~\ref{prop:multivariate} and \ref{prop:multivariate-bis} are in fact equivalent to (see similar proof in \cite{self}):
\BEQ
\label{eq:FFF}
\forall u,v,w \in \rb^p, \ |F'''[u,v,t]| \leqslant \betac \| u\|_2 [ v^\top F''(w)v ]^{1/2} [ t^\top F''(w) t]^{1/2}.
\EEQ

\subsection{Univariate functions}

We first consider univariate functions and prove the following lemma that gives upper and lower Taylor expansions:
\begin{lemma}
\label{lemma:univariate}
Let $g$ be  a convex three times differentiable   function $g: \rb \mapsto \rb$ such that  for all $t \in \rb$, $| g'''(t) | \leqslant \betacuni g''(t)$,
for some $\betacuni \geqslant 0$. Then, for all $t \geqslant 0$:
%\BEQ
 %g''(0) e^{-\betacuni t} \leqslant    g''(t)   \leqslant  g''(0) e^{\betacuni t} ,
%\EEQ
\BEQ
\label{eq:14}
 \frac{g''(0)}{  \betacuni^{2}} (e^{-\betacuni t} +  \betacuni t - 1)
 \leqslant  g(t)  - g(0) -   g'(0) t \leqslant
  \frac{g''(0)}{  \betacuni^{2}} (e^{\betacuni t} -  \betacuni t - 1).
\EEQ
\end{lemma}
\begin{proof} Let us first assume that $g''(t)$ is strictly positive for all $t \in \rb$.
We have, for all $t \geqslant 0$:
 $
-\betacuni \leqslant \frac{d \log g''(t)}{dt} \leqslant \betacuni$. 
Then, by integrating once  between $0$ and $t$, taking exponentials, and then integrating twice:
$$
-\betacuni t \leqslant   \log g''(t) - \log g''(0)  \leqslant \betacuni t,
$$
\BEQ
\label{eq:qq}
 g''(0) e^{-\betacuni t} \leqslant    g''(t)   \leqslant  g''(0) e^{\betacuni t}, 
\EEQ
$$
g''(0)    \betacuni^{-1}(1 - e^{-\betacuni t})\leqslant    g'(t) - g'(0)   \leqslant  g''(0)  \betacuni^{-1}(e^{\betacuni t} - 1),
$$
\BEQ
\label{eq:lowerboundg}
 g(t)  \geqslant g(0) +   g'(0)t + g''(0)  \betacuni^{-2}(e^{-\betacuni t} +  \betacuni t - 1),
\EEQ
\BEQ
\label{eq:upperboundg}
 g(t)  \leqslant g(0) +  g'(0) t+ g''(0)  \betacuni^{-2}(e^{\betacuni t} - \betacuni t - 1),
\EEQ
which leads to \eq{14}.

Let us now assume only that $g''(0)>0$. If we denote by $A$ the connected component that contains 0 of the open set $\{ t\in \rb, \ g''(t)>0\}$, then the preceding developments are valid on $A$; thus, \eq{qq} implies that $A$ is not upper-bounded. The same reasoning on $-g$ ensures that $A = \rb$ and hence $g''(t)$ is strictly positive for all $t \in \rb$. 
Since the problem is invariant by translation, we have shown that if there exists $t_0 \in \rb$ such that $g''(t_0)>0$, then for all $t \in \rb$, $g''(t)>0$.

Thus, we need to prove \eq{14} for $g''$ always strictly positive (which is done above) and for $g''$ identically equal to zero, which implies that $g$ is linear, which is then equivalent to   \eq{14}.
\end{proof}
Note the difference with  a classical uniform bound on the third derivative, which leads to a third-order polynomial   lower bound, which tends to $-\infty$ more quickly than \eq{lowerboundg}.
Moreover, \eq{upperboundg} may be interpreted as an upperbound on the remainder in the Taylor expansion of $g$ around $0$:
$$g(t) - g(0) -  g'(0)t - \frac{g''(0)}{2} t^2
\leqslant  g''(0)  \betacuni^{-2}(e^{\betacuni t} -  \frac{1}{2}\betacuni^2 t^2 - \betacuni t - 1).
 $$
The right hand-side is equivalent to $\frac{\betacuni t^3 }{6} g''(0)$ for $t$ close to zero (which should be expected from a three-times differentiable function such that $g'''(0) \leqslant S g''(0)$), but still provides a good bound for $t$ away from zero (which cannot be obtained from a regular Taylor expansion).

Throughout the proofs, we will use the fact that the functions $u \mapsto \frac{e^{u}-1}{u}$ and $u \mapsto  \frac{e^{u}-1 - u}{u^2}$ can be extended to continuous functions on $\rb$, which are thus bounded on any compact. The bound will depend on the compact and can be obtained easily.
 
\subsection{Proof of Proposition~\ref{prop:multivariate}}
\label{app:multivariate}
By applying Lemma~\ref{lemma:univariate} (\eq{lowerboundg} and \eq{upperboundg}) to $g(t) = F(w+tv)$ (with constant $\betacuni = \betac \| v\|_2$) and taking $t=1$, we get the desired first two inequalities in \eq{lowerboundF} and \eq{upperboundF}.  By considering the function $g(t) = u^\top F''(w + t v) u$, we have
  $g'(t) = F''' (w+tv) [ u,u,v]$, which is such that
  $|g'(t)| \leqslant \| v \|_2 \betac g(t)$, leading to $ g(0) e^{ -\| v \|_2 \betac t} \leqslant g(t) \leqslant g(0) e^{ \| v \|_2 \betac t}$, and thus to \eq{boundhessianF} for
  $t=1$ (when considered for all $u\in \rb^p$).
  
   In order to prove \eq{boundgradientF}, we consider $h(t) = z ^\top ( F'(w+tv) - F'(w) -   F''(w)v t )$. We have $h(0)=0$, $h'(0)=0$ and
  $h''(t) = F'''(w+tv)[v,v,z] 
\leqslant  \betac \| v\|_2 e^{t \betac \| v\|_2} [  z^\top F''(w) z]^{1/2} [  v^\top F''(w) v ]^{1/2}
  $ using \eq{boundhessianF} and \eq{FFF}. Thus, by integrating between $0$ and $t$, 
  $$h'(t) \leqslant  [  z^\top F''(w) z]^{1/2} [  v^\top F''(w) v ]^{1/2}  (e^{ t \betac \| v\|_2} - 1),$$ which implies
  $\textstyle h(1) \leqslant [  z^\top F''(w) z]^{1/2} [  v^\top F''(w) v]^{1/2} \int_0^1  (e^{ t \betac \| v\|_2} - 1)  dt,$
  which in turn leads to \eq{boundgradientF}.

Using similar techniques, i.e., by considering the function  $t \mapsto
= z^\top [ F''(w+tv) - F''(w) ]u $,
we can prove that for all $z,u,v,w \in \rb^p$, we have:
\BEQ
\label{eq:new}
z^\top [ F''(w+v) - F''(w) ]u 
\leqslant \frac{ e^{ \betac \| v\|_2 } - 1}{\| v\|_2 } [ v^\top F''(w) v ]^{1/2}
[ z^\top F''(w) z ]^{1/2} \| u\|_2.
\EEQ

\subsection{Proof of Proposition~\ref{prop:multivariate-bis}}
Since we have assumed that $\lambda(w)>0$,  then by \eq{boundhessianF}, the Hessian of $F$ is everywhere invertible, and hence the function $F$ is strictly convex. Therefore,
if the minimum is attained, it is unique.

Let $v \in \rb^p $ be such that $ v^\top F''(w) v = 1$. Without loss of generality, we may assume that $F'(w)^\top v$ is negative. This implies that for all $t \leqslant 0$,
$F(w+tv) \geqslant F(w)$. Moreover, let us denote
$\kappa = - v^\top F'(w) \betac \| v \|_2$, which is nonnegative and such that
$\kappa   
\leqslant
\frac{\betac  | v^\top F'(w)  |    }{\lambda(w)^{1/2}}  
\leqslant  \frac{\betac    \nu(F,w)       }{\lambda(w)^{1/2}} \leqslant 1/2$.
 From \eq{lowerboundF}, for all $t  \geqslant 0 $, we have:
 \BEAS
   F(w+tv) & \geqslant &  F(w)  +   v^\top F'(w) t     +  \frac{ 1}{\betac^2 \| v \|_2^2   }
(e^{-\betac \| v \|_2 t } + \betac \| v \|_2 t - 1)  \\
    & \geqslant &  F(w)  
   +  \frac{ 1}{\betac^2 \| v \|_2^2   } 
   \left[
   e^{-\betac \| v \|_2 t } +  ( 1- \kappa) \betac \| v \|_2 t - 1
   \right].\EEAS
Moreover,  a short calculation shows that for all $ \kappa \in (0,1] $:
\BEQ
\label{eq:kappa}
 e^{- 2 \kappa ( 1- \kappa)^{-1}} +  ( 1- \kappa) 2 \kappa ( 1- \kappa)^{-1} - 1 \geqslant 0.
 \EEQ
This implies that for $t_0= 2 (\betac \| v \|_2)^{-1} \kappa ( 1- \kappa)^{-1}$,
$F(w+t_0v) \geqslant  F(w)$. 
Since $t_0 \leqslant   \frac{2}{1-\kappa}| v^\top F'(w)|
\leqslant  {2}{ \nu(F,w)}\left(
 1 -  \frac{\nu(F,w) \betac}{ \lambda(w)^{1/2} } \right)^{-1} \leqslant 4 \nu(F,w)$, we have $F(w+tv) \geqslant F(w)$ for $ t= 4\nu(F,w)$.

Since this is true for all $v$ such that $ v^\top F''(w) v = 1$, this shows that the value of the function $F$ on the entire ellipsoid (since $F''(w)$ is positive definite) $v^\top F''(w) v = 16 \nu(F,w)^2$ is greater or equal to the value at $w$; thus, by convexity, there must be a minimizer $w^\ast$---which is unique because of \eq{boundhessianF}---of $F$ such that
$$
 (w - w^\ast)^\top{F''(w)} (w - w^\ast)  \leqslant 
 {16}{ \nu(F,w)^2},$$
leading to \eq{nu2}. 

In order to prove \eq{nu4}, we will simply apply \eq{nu2} at $w+v$, which requires to upper-bound $\nu(F,w+v)$.
  If we denote by $v = - F''(w) ^{-1} F'(w)$  the Newton step, we have:
  \BEAS
  & & \| F''(w) ^{-1/2}  F'(w+v)\|_2 \\
  & = & \big\| F''(w) ^{-1/2}  [ F'(w+v) - F'(w) - F''(w) v ]
  \big\|_2 \\
  & = & \bigg\| \int_{0}^1 F''(w) ^{-1/2}  [ F''(w+tv) -
  F''(w) ] v dt
  \bigg\|_2 \\
   & \leqslant &  \int_{0}^1  \bigg\|F''(w) ^{-1/2}  [ F''(w+tv) -
  F''(w) ] F''(w) ^{-1/2} F''(w) ^{1/2}v  \bigg\|_2  dt
  \\
  & \leqslant &  \int_{0}^1  \bigg\| \left[ F''(w) ^{-1/2}   F''(w+tv)  F''(w) ^{-1/2} -
  \idm \right]  F''(w) ^{1/2}v  \bigg\|_2  dt.
  \EEAS
  Moreover, we have from \eq{boundhessianF}:
  $$
  (e^{-t \betac \| v \|_2 } - 1 ) \idm   \preccurlyeq F''(w) ^{-1/2}   F''(w+tv)  F''(w) ^{-1/2} -
  \idm \preccurlyeq  (e^{t \betac \| v \|_2 } - 1 ) \idm.
  $$
  Thus,
   \BEAS
  \| F''(w) ^{-1/2}  F'(w+v)\|_2
  & \!\!\!\! \leqslant \!\!\!\! &
   \int_{0}^1  \! \max\{ e^{t \betac \| v \|_2 } - 1 , 1 - e^{-t \betac \| v \|_2 } \}    \|F''(w) ^{1/2}v\|_2   dt  \\
   & \!\!\!\!=\!\!\!\! & \nu(F,w)  \int_{0}^1\!\! ( e^{t \betac \| v \|_2 }\! -\! 1  )  dt 
   = \nu(F,w) \frac{e^{ \betac \| v \|_2 } \!-\! 1 \! -\! \betac \| v \|_2}{\betac \| v \|_2 } .\EEAS
  Therefore, using \eq{boundhessianF} again, we obtain:
  $$
  \nu(F,w+v) = \| F''(w+v) ^{-1/2}  F'(w+v)\|_2  \leqslant \nu(F,w) e^{ \betac \| v \|_2 /2} \frac{e^{ \betac \| v \|_2 } - 1  - \betac \| v \|_2}{\betac \| v \|_2 } .
  $$
  We have $\betac \| v \|_2 \leqslant \betac \lambda^{-1/2}  \nu(F,w) \leqslant 1/2$, and thus, we have  $$e^{ \betac \| v \|_2 /2} \frac{e^{ \betac \| v \|_2 } - 1  - \betac \| v \|_2}{\betac \| v \|_2 }  \leqslant   \betac \| v \|_2 \leqslant R\nu(F,w) \lambda(w)^{-1/2},$$ leading to:
\BEQ
\label{eq:nuwv}
\nu(F,w+v) \leqslant \frac{\betac}{\lambda(w)^{1/2}}\nu(F,w)^2.
\EEQ
Moreover, we have:
  \BEAS
\frac{  \betac  \nu(F,w+v)}
{\lambda(w+v)^{1/2} }
& \!\!\leqslant\!\!&  \frac{\betac e^{\betac \| v \|_2 / 2} }{\lambda(w)^{1/2}}
\nu(F,w+v)
\leqslant  
 \frac{\betac  }{\lambda(w)^{1/2}}
  \nu(F,w)   e^{ \betac \| v \|_2 } \frac{e^{ \betac \| v \|_2 } \!-\! 1\!  -\! \betac \| v \|_2}{\betac \| v \|_2 } , \\
  & \!\!\leqslant \!\!&  \frac{\betac  }{\lambda(w)^{1/2}}
  \nu(F,w)  \times \betac \| v \|_2 \leqslant 
  \bigg(  \frac{\betac  }{\lambda(w)^{1/2}}
  \nu(F,w) \bigg)^2 \leqslant 1/4,
\EEAS
which leads to \eq{nunew}. Moreover, it shows that we can apply \eq{nu2} at $w+v$ and get:
\BEAS
& &[ ( w^\ast - w - v)^\top {F''(w)}  ( w^\ast - w - v)]^{1/2}  \\
& \leqslant & e^{ \betac \| v \|_2/2 }  [ ( w^\ast - w - v)^\top {F''(w+v)}  ( w^\ast - w - v)]^{1/2}  \\
& \leqslant & 4e^{ \betac \| v \|_2/2 }   \nu(F,w+v) \leqslant 4 \betac \| v \|_2  \nu(F,w),
\EEAS
which leads to the desired result, i.e., \eq{nu4}.
 
\section{Proof of Theorem~\ref{theo:mis}}
\label{app:mis}
Following \cite{sridharan2008fast,steinwart2006new}, we denote by $w_\lambda$ the unique global minimizer of the expected regularized risk
$J_\lambda(w) = J_0(w) + \frac{\lambda}{2} \|w\|_2^2$.
We simply apply \eq{nu2} from Proposition~\ref{prop:multivariate-bis} to $\hat{J}_\lambda$ and $w_\lambda$, to obtain, if the Newton decrement (see \mysec{self} for its definition) $\nu(\hat{J}_\lambda,w_\lambda)^2$ is less than $\lambda /4R^2$, 
that $\hat{w}_\lambda$ and its population counterpart $w_\lambda$ are close, i.e.:
$$
(\hat{w}_\lambda - w_\lambda)^\top \hat{J}_\lambda''(w_\lambda)
(\hat{w}_\lambda - w_\lambda) \leqslant 16 \nu(\hat{J}_\lambda,w_\lambda)^2.
$$
We can then apply the upper Taylor expansion  in \eq{upperboundF} from Proposition~\ref{prop:multivariate} to $J_\lambda$ and $w_\lambda$, to obtain, with $v = \hat{w}_\lambda - w_\lambda$ (which is such that
$R \| v\|_2 \leqslant 4 \frac{ R \nu(\hat{J}_\lambda,w_\lambda)}{\lambda^{1/2}}
\leqslant 2$):
$$J_\lambda(\hat{w}_\lambda) -  J_\lambda(w_\lambda)  \leqslant
 \frac{ v^\top J_\lambda''(w_\lambda)  v}{R^2 \| v \|_2^2  }
(e^{R \| v \|_2  } - R \| v \|_2   - 1)
\leqslant 20 \nu(\hat{J}_\lambda,w_\lambda)^2. $$
Therefore, for any $w_0 \in \rb^p$, since $w_\lambda$ is the minimizer
of $J_\lambda(w) = J_0(w) + \frac{\lambda}{2} \| w\|_2^2$:
\BEQ
\label{eq:H}
J_0(\hat{w}_\lambda) \leqslant  J_0(w_0) + \frac{\lambda}{2} \| w_0\|_2^2
+ 20 \nu(\hat{J}_\lambda,w_\lambda)^2.
\EEQ
We can now apply the concentration inequality from Proposition~\ref{prop:concentration} in Appendix~\ref{app:concentration}, i.e.,
\eq{concentration-no-assumptions}, with $u = \log(8/\delta)$. We use 
$\lambda =  { 19R^2} \sqrt{\frac{\log(8/\delta)}{n}}$. In order to actually have
$\nu(\hat{J}_\lambda,w_\lambda) \leqslant \lambda^{1/2}/2R$ (so that we can apply our self-concordant analysis), it is sufficient that:
$$
41 R^2 u / \lambda n \leqslant \lambda / 8 R^2
, \ 
63  (u/n)^{3/2}   R^2  / \lambda  \leqslant \lambda / 16 R^2
, \ 
8 (u/n)^{ 2}  R^2  / \lambda   \leqslant \lambda / 16 R^2,
$$
leading to the constraints $u \leqslant n / 125$. We then get with probability at least $1-\delta = 1 - 8 e^{-u}$ (for $u \leqslant n / 125$):
%$$
%\nu(\hat{J}_\lambda,w_\lambda)^2 \leqslant \sqrt{\frac{u}{n}}
%\times \left(
%41/19 + 63/19 \times 125^{-1/2} + 8/19 \times 125^{-3/2} 
%\right) \leqslant 3\sqrt{\frac{u}{n}}.
%$$
%This implies
$$
J_0(\hat{w}_\lambda) \leqslant  J_0(w_0) + \frac{\lambda}{2} \| w_0\|_2^2
+ 20  \frac{\lambda}{4 R^2 }
\leqslant   J_0(w_0)  + \frac{ ( 10 + 100 R^2 \|w_0\|_2^2) \sqrt{\log(8/\delta)}}{\sqrt{n}}.
$$
For $u > n/125$, the bound in \eq{mis} is always satisfied. Indeed, this implies with our choice of $\lambda$ that $\lambda \geqslant  R^2$. Moreover, since
  $\|\hat{w}_\lambda\|_2^2$ is bounded from above by $\log(2) \lambda^{-1} \leqslant R^{-2}$,
$$
J_0(\hat{w}_\lambda) \leqslant J_0(w_0) + \frac{R^2}{2} \| \hat{w}_\lambda - w_0\|_F^2 \leqslant  J_0(w_0) + 1 + R^2 \| w_0\|_2^2,
$$
which is smaller than the right hand-side of \eq{mis}.

\section{Proof of Theorem~\ref{theo:perf}}
\label{app:perf}

 We   denote  by $J_0^T$ the second-order Taylor expansion of $J_0$ around $w_0$, equal to $J_0^T(w) = J_0(w_0) + \frac{1}{2} ( w-w_0)^\top  Q ( w - w_0)$, with $Q = J_0''(w_0)$, and $\hat{J}_0^T$ the expansion of $\hat{J}_0$ around $w_0$, equal to
  $J_0^T(w) - q^\top  w $. 
  We denote by $\hat{w}_\lambda^N$ the one-step Newton iterate from $w_0$ for the function $\hat{J}_0$, defined as the global minimizer of $\hat{J}_0^T$ and equal to
  $\hat{w}_\lambda^N = w_0 + ( Q +\lambda \idm)^{-1} ( q - \lambda w_0)$.
  
  What the following proposition shows is that we can replace $\hat{J}_0$ by $\hat{J}_0^T$ for obtaining the estimator and that we can replace $J_0$ by $J_0^T$ for measuring its performance, i.e., we may do as if we had a weighted least-squares cost, as long as the Newton decrement is small enough:

\begin{proposition}[Quadratic approximation of risks]
\label{prop:taylor}
Assume $ \nu( \hat{J}_\lambda,w_0)^2 = ( q - \lambda w_0)^\top ( Q + \lambda \idm)^{-1} ( q - \lambda w_0)  \leqslant \frac{\lambda}{4R^2} $. We have:
\BEQ
\label{eq:taylor}
| J_0(\hat{w}_\lambda) - J_0^T(\hat{w}_\lambda^N) |
\leqslant 
 \frac{  15 R \nu(\hat{J}_\lambda, w_0)^2}{ \lambda^{1/2}} \| Q^{1/2}( \hat{w}_\lambda^N - w_0 )\|_2
+  \frac{ 40 R^2}{\lambda}  \nu(\hat{J}_\lambda,w_0)^4.
\EEQ
\end{proposition}
\begin{proof}
We show that (1) $\hat{w}_\lambda^N$ is close to $\hat{w}_\lambda$ using Proposition~\ref{prop:multivariate-bis} on the behavior of Newton's method, (2) that
 $\hat{w}_\lambda^N$ is close to $w_0$ by using its closed form
 $\hat{w}_\lambda^N = w_0 + ( Q+ \lambda \idm)^{-1}( q - \lambda w_0)$, and (3) that $J_0$ and $J_0^T$ are close  using Proposition~\ref{prop:multivariate} on upper and lower Taylor expansions.

 We first apply \eq{nu4} from Proposition~\ref{prop:multivariate-bis} to get \BEQ
\label{eq:B}
(\hat{w}_\lambda - \hat{w}_\lambda^N)^\top     \hat{J}_\lambda''(w_0) (\hat{w}_\lambda - \hat{w}_\lambda^N) \leqslant 
 \frac{ 16 R^2}{\lambda} \nu(\hat{J}_\lambda, w_0)^4.
\EEQ
This implies that $\hat{w}_\lambda$ and $\hat{w}_\lambda^N$ are close, i.e.,
\BEAS
\| \hat{w}_\lambda -\hat{w}_\lambda^N \|^2 & \leqslant &  \lambda^{-1} 
(\hat{w}_\lambda -\hat{w}_\lambda^N )^\top  \hat{J}_\lambda''(w_0) ( \hat{w}_\lambda -\hat{w}_\lambda^N  ) \\
& \leqslant &    \frac{ 16 R^2}{\lambda^2} \nu(\hat{J}_\lambda, w_0)^4
\leqslant 
 \frac{ 4}{\lambda} \nu(\hat{J}_\lambda, w_0)^2 \leqslant \frac{1}{R^2}.
\EEAS
Thus, using the closed form expression for $\hat{w}_\lambda^N
= w_0 + ( Q +\lambda \idm)^{-1} ( q - \lambda w_0)$, we obtain
\BEAS
\| \hat{w}_\lambda - w_0 \| & \leqslant &
\| \hat{w}_\lambda -\hat{w}_\lambda^N \| + 
\| w_0 - \hat{w}_\lambda^N \|
 \\
 & \leqslant &  2\frac{ \nu(\hat{J}_\lambda, w_0) }{  \lambda^{1/2}}
 + \frac{ \nu(\hat{J}_\lambda, w_0)}{ \lambda^{1/2}} 
\leqslant  \frac{ 3 \nu(\hat{J}_\lambda, w_0)}{ \lambda^{1/2}}
\leqslant \frac{3}{2R}.
\EEAS
We can now apply \eq{lowerboundF} from Proposition~\ref{prop:multivariate-bis} to get for all $v$ such that 
$R \| v\|_2 \leqslant 3/2$, 
\BEQ
\label{eq:ASD}
| J_0(w_0+v) - J_0^T(w_0+v) | \leqslant ( v^\top Q v ) R \| v\|_2 /4.
\EEQ
 Thus, using \eq{ASD} for $v = \hat{w}_\lambda - w_0$ and $v = \hat{w}_\lambda^N - w_0$ :
\BEAS
& & | J_0(\hat{w}_\lambda) - J_0^T(\hat{w}_\lambda^N) | \\
&\!\!\! \!\!\! \leqslant \!\!\! \!\!& 
| J_0(\hat{w}_\lambda) - J_0^T(\hat{w}_\lambda)| + 
| J_0^T(\hat{w}_\lambda^N) - J_0^T(\hat{w}_\lambda)|, \\
 & \!\!\! \!\!\!\leqslant \!\!\!&  
\frac{R}{4}  \| \hat{w}_\lambda - w_0 \|_2 \  \| Q^{1/2}( \hat{w}_\lambda - w_0 )\|_{2}^2
+ \frac{1}{2} \left| \| Q^{1/2}(\hat{w}_\lambda - w_0 )\|_{2}^2
 - \|  Q^{1/2}(\hat{w}_\lambda^N - w_0 )\|_{2}^2 \right|,
\\
&\!\!\! \! \!\!\leqslant\! \!\! \!\!&  
\frac{ 3 R \nu(\hat{J}_\lambda, w_0)}{ 4 \lambda^{1/2}} \| Q^{1/2}( \hat{w}_\lambda - w_0 )\|_{2}^2
+ \frac{1}{2} \left| \| Q^{1/2}( \hat{w}_\lambda - w_0) \|_{2}^2
 - \|  Q^{1/2}(\hat{w}_\lambda^N - w_0) \|_{2}^2 \right|,
\\
&\!\!\! \! \!\!\leqslant\!\!\!\!\! &  
\frac{ 3 R \nu(\hat{J}_\lambda, w_0)}{ 4 \lambda^{1/2}} \|  Q^{1/2}(\hat{w}_\lambda^N - w_0 )\|_{2}^2
+ {\textstyle \big( \frac{1}{2}\!+\! \frac{3}{4} \big) }\left| \| Q^{1/2}( \hat{w}_\lambda - w_0 )\|_{2}^2
 - \| Q^{1/2}( \hat{w}_\lambda^N - w_0 )\|_{2}^2 \right|,
\\
&\!\!\! \! \!\!\leqslant\!\!\! \!\!&  
\frac{ 3 R \nu(\hat{J}_\lambda, w_0)}{ 4 \lambda^{1/2}} \|  Q^{1/2}(\hat{w}_\lambda^N - w_0) \|_{2}^2 \\
& & 
+   \frac{5}{4} \|  Q^{1/2}(\hat{w}_\lambda - \hat{w}_\lambda^N)  \|_{2}^2
+ \frac{5}{2} \|  Q^{1/2}(\hat{w}_\lambda - \hat{w}_\lambda^N ) \|_{2}
\|  Q^{1/2}(\hat{w}_\lambda^N - w_0) \|_{2}.
 \EEAS

From \eq{B}, we have
$
 \| Q^{1/2}( \hat{w}_\lambda - \hat{w}_\lambda^N)  \|_{2}^2  \leqslant \frac{ 16 R^2}{\lambda}  \nu(\hat{J}_\lambda,w_0)^4
$.
We thus obtain, using that $\| Q^{1/2}( \hat{w}_\lambda^N - w_0 )\|_{2} \leqslant \nu(\hat{J}_0,w_0)$:
$$
| J_0(\hat{w}_\lambda) - J_0^T(\hat{w}_\lambda^N) |
\!\!\leqslant \!\!
\left(
\frac{3}{4} \!+ \!\frac{5}{2} \sqrt{32}
\right)\frac{   \nu(\hat{J}_\lambda, w_0)^2}{R^{-1} \lambda^{1/2}} \|  Q^{1/2}(\hat{w}_\lambda^N - w_0) \|_{2}
+  \frac{ 40 R^2}{\lambda}  \nu(\hat{J}_\lambda,w_0)^4,
$$
which leads to the desired result.
\end{proof}

We can now go on with the proof of Theorem~\ref{theo:perf}. From \eq{taylor} in Proposition~\ref{prop:taylor} above, we have,
if $\nu(\hat{J}_\lambda, w_0)^2 \leqslant \lambda / 4R^2$,
\BEAS
J_0(\hat{w}_\lambda) &  = &  J_0^T(\hat{w}_\lambda^N) + B  \\
& = &  J_0(w_0) + \frac{1}{2}(q - \lambda w_0)^\top Q ( Q + \lambda \idm)^{-2} ( q - \lambda w_0 ) + B \\
& = &    J_0(w_0) + \frac{d_2}{2n} +\frac{b_2}{2} + B + C, 
\EEAS
\BEAS
\mbox{ with } C &  = &   {\lambda}  w_0^\top   (Q + \lambda \idm)^{-2}  Q q  
+    \frac{1}{2}
 \tr  (Q + \lambda \idm)^{-2} Q \left( qq^\top \! - \! \frac{1}{n} Q \right) , \\
| B |  & \leqslant &   \frac{  15 R \nu(\hat{J}_\lambda, w_0)^2}{ \lambda^{1/2}} \| Q^{1/2}( \hat{w}_\lambda^N - w_0 )\|_2
+  \frac{ 40 R^2}{\lambda}  \nu(\hat{J}_\lambda,w_0)^4.
\EEAS
We can now bound each term separately and check that we indeed have $\nu(\hat{J}_\lambda,w_0)^2 \leqslant \lambda / 4R^2$ (which allows to apply Proposition~\ref{prop:multivariate-bis}).
First, from \eq{kappa2}, we can derive
$$
b_2+\frac{d_2}{n} \leqslant  b_1+\frac{d_1}{n} \leqslant \frac{\kappa \lambda^{1/2}}{R} \big( b_2+\frac{d_2}{n} \big)^{1/2} 
 \leqslant \frac{\kappa \lambda^{1/2}}{R} \big( b_1+\frac{d_1}{n} \big)^{1/2}, 
$$
which implies the following identities:
\BEQ
\label{eq:b1b2}
b_2+\frac{d_2}{n} \leqslant  b_1+\frac{d_1}{n} \leqslant   
 \frac{\kappa^2 \lambda}{R^2} .
\EEQ
We have moreover:
\BEAS
\nu(\hat{J}_\lambda,w_0)^2  & = &  ( q    - \lambda w_0)^\top (  Q  + \lambda \idm)^{-1} ( q    - \lambda w_0) 
\\
&\leqslant & b_1 + \frac{d_1}{n} + \tr  \big(  Q  + \lambda \idm)^{-1} \bigg( qq^\top - \frac{Q}{n} \bigg) +   {2\lambda}  w_0^\top   (Q + \lambda \idm)^{-1}  q .
\EEAS
We can now apply concentration inequalities from Appendix~\ref{app:concentration}, together with the following applications of Bernstein's inequality. Indeed,
we have
$
 {\lambda}{ }  w_0^\top   (Q + \lambda \idm)^{-2}  Q q 
= \sum_{i=1}^n Z_i,
$
with 
\BEAS
|Z_i| &  \leqslant &  \frac{\lambda}{n} | w_0^\top   (Q + \lambda \idm)^{-2}  Q x_i|  \\
& \leqslant & 
\frac{\lambda}{2n} \left(
w_0^\top   (Q + \lambda \idm)^{-2}  Q w_0
\right)^{1/2}
\left(
x_i^\top   (Q + \lambda \idm)^{-2}  Q x_i
\right)^{1/2}  \leqslant 
\frac{ b_2^{1/2} }{2n}   R \lambda^{-1/2}
.
\EEAS
Moreover, $ \E Z_i^2    \leqslant   
\frac{ \lambda^2}{n} w_0^\top  (Q + \lambda \idm)^{-2}  Q^3  (Q + \lambda \idm)^{-2} w_0 \leqslant \frac{1}{n} b_2$.
We can now apply Bernstein inequality~\cite{massart-concentration} to get with probability at least $1-2e^{-u}$ (and using \eq{b1b2}):
$$
 {\lambda}{ }  w_0^\top   (Q + \lambda \idm)^{-2}  Q q
\leqslant \sqrt{ \frac{ 2 b_2 u }{n}} + \frac{u}{6 n}
 b_2^{1/2}   R \lambda^{-1/2}
 \leqslant \sqrt{ \frac{ 2 b_2 u }{n}} + \frac{u \kappa}{6 n}
 .
 %\leqslant 2 \sqrt{ \frac{b_2 u}{n} }  .
$$
Similarly, with probability at least
$1-2e^{-u}$, we have:
$$
 {\lambda}{ }  w_0^\top   (Q + \lambda \idm)^{-1}  q
\leqslant \sqrt{ \frac{2 b_2 u }{n}} + \frac{u\kappa}{6 n}
.
 %\leqslant 2 \sqrt{ \frac{b_2 u}{n} }  .
$$
We thus get, through the union bound, with probability at least $1 - 20 e^{-u}$:
\BEAS
\nu(\hat{J}_\lambda,w_0)^2 &  \leqslant & 
\bigg( b_1 + \frac{d_1}{n} \bigg)
+ \bigg( \frac{32 d_2^{1/2} u^{1/2}}{n} + \frac{18u}{n} + \frac{ 53 R d_1^{1/2} u^{3/2}}{n^{3/2}\lambda^{1/2}}  + 9 \frac{R^2 u^2}{\lambda n^2} \bigg)\\
& & \hspace*{6.5cm}  + 
 \bigg( 2\sqrt{ \frac{2 b_2 u }{n}} + \frac{\kappa u}{6 n}
  \bigg),
\\
 &  \leqslant & 
b_1 + \frac{d_1}{n} 
+ \frac{64 u^{1/2}}{n^{1/2}}  \big( b_2 + \frac{d_2}{n} \big)^{1/2} + \frac{u}{n} \big( 18+\frac{\kappa}{6} \big)  + 
 \frac{R^2  }{\lambda }  \frac{ 9 u^2  }{n^2  }   \\
& &  \hspace*{7cm} + \frac{ 53  n^{1/2} \kappa u^{3/2}}{n^{3/2} },  \\
 & \leqslant & \frac{\lambda \kappa^2}{R^2} + E,
 \EEAS
 together with $C \leqslant E$.
 We now take $u = (nb_2 + d_2) v^2$ and assume $v\leqslant 1/4$, $\kappa \leqslant 1/16$, and $ v^3 ( nb_2 + d_2)^{1/2} \leqslant 12$, so that, we have
 \BEAS
 E & \leqslant & 
  {64 v} \big( b_2 + \frac{d_2}{n} \big) + v^2 \big( b_2 + \frac{d_2}{n} \big) \big( 18+\frac{\kappa}{6} \big)  + 
 \frac{9 R^2  }{\lambda }  v^4 \big( b_2 + \frac{d_2}{n} \big)^2  \\
 & & \hspace*{7cm}  +  { 53  n^{1/2} \kappa }
 v^3 \big( b_2 + \frac{d_2}{n} \big)^{3/2},   \\
 & \leqslant &  
  \big( b_2 + \frac{d_2}{n} \big)
  \bigg( 64 v + \big( 18+\frac{\kappa}{6} \big)  v^2 + 
   \frac{9 R^2  }{\lambda }  v^4  \frac{ \lambda \kappa^2}{R^2}
   + 53 \kappa v^3 ( nb_2 + d_2)^{1/2}
  \bigg),
  \\
 & \leqslant &  
  \big( b_2 + \frac{d_2}{n} \big)
  \bigg( 64 v + 18 v^2 +\frac{\kappa}{6}   v^2 + 9 \kappa^2 v^4
   + 53 \kappa v^3 ( nb_2 + d_2)^{1/2}
  \bigg),
  \\
   & \leqslant &  
  \big( b_2 + \frac{d_2}{n} \big)
  \bigg( 68.5 v  + \frac{\kappa}{6 \times 16}  + 9 \kappa / 16 \times 16 \times 16   + 53 \kappa \times \frac{ 12}{64}
  \bigg),
  \\
 & \leqslant &  
  \big( b_2 + \frac{d_2}{n} \big)
  \big( 69 v  +  10 \kappa
  \big) \leqslant 20 \big( b_2 + \frac{d_2}{n} \big)
   .
  \EEAS
 This implies that $\nu(\hat{J}_\lambda,w_0)^2 \leqslant \frac{\lambda}{R^2} \frac{20}{256} \leqslant \frac{\lambda}{4 R^2} $, so that we can apply Proposition~\ref{prop:multivariate-bis}.   
  Thus, by denoting $e_2 = b_2 + \frac{d_2}{n}$, $e_1 = b_1 + \frac{d_1}{n}$, and
  $\alpha =   69 v  +  10 \kappa
 \leqslant 20$, we get a global upper bound:
$$B + |C| \leqslant e_2 \alpha + \frac{40R^2}{\lambda} ( e_1 + e_2 \alpha )^2 + \frac{15Re_2^{1/2}}{\lambda^{1/2}}
( e_1  + e_2 \alpha ) ( 1 + \alpha )^{1/2} .
$$
With $e_1 + e_2 \alpha \leqslant e_2^{1/2}  ( \kappa \lambda^{1/2}/R ) ( 1 + \alpha)$, we get
\BEAS
B+ |C| & \leqslant & e_2 \alpha +  40 \kappa^2  e_2 ( 1 + \alpha)^2 + 15 \kappa e_2  ( 1 + \alpha)^{3/2} \\
&  \leqslant&  e_2 \alpha + e_2 \kappa ( 40 \times 21 \times 21 / 16 + 15  ( 21)^{3/2} )
\leqslant e_2 ( 69 v + 2560 \kappa),
\EEAS
which leads to the desired result, i.e., \eq{perf}.

\section{Proof of Theorem~\ref{theo:smooth}}
\label{app:smooth}
We follow the same proof technique than for Theorem~\ref{theo:perf} in Appendix~\ref{app:perf}.
We have:
\BEAS
\!\! J_0(\hat{w}_\lambda) & \!\!\!\!=\!\! \!\!&  \hat{J}_0(\hat{w}_\lambda) + q^\top( \hat{w}_\lambda - w_0 ) + q^\top w_0 \\
  & \!\!\!\!=\!\! \!\!&  \hat{J}_0(\hat{w}_\lambda)  + q^\top( \hat{w}_\lambda - \hat{w}_\lambda^{NN} )    + q^\top( \hat{w}_\lambda^N - w_0 ) 
  - q^\top \hat{J}_\lambda''(\hat{w}^N_\lambda)^{-1}\hat{J}_\lambda'(\hat{w}^N_\lambda) + q^\top w_0 ,   \EEAS
 where $\hat{w}_\lambda^{NN}$ is the two-step Newton iterate from $w_0$. We have, from \eq{nuwv}, 
$\nu(\hat{J}_\lambda,\hat{w}_\lambda^N) \leqslant \frac{2\betac}{\lambda^{1/2}}\nu(\hat{J}_\lambda,w_0)^2$, which then implies (with \eq{nu4}):
 $$
 (\hat{w}_\lambda - \hat{w}_\lambda^{NN})^\top (Q+\lambda \idm) (\hat{w}_\lambda - \hat{w}_\lambda^{NN} ) \leqslant 
\frac{16 R^2}{\lambda}  \bigg( \frac{2\betac}{\lambda^{1/2}}\nu(\hat{J}_\lambda,w_0)^2 \bigg)^4
 \leqslant
 \frac{512 R^6 \nu(\hat{J}_\lambda,w_0)^8}{\lambda^3}
, $$
 which in turn implies
 \BEA
 \nonumber
 |
  q^\top( \hat{w}_\lambda - \hat{w}_\lambda^{NN} ) |
  & \leqslant &  [ q(Q+\lambda \idm)^{-1} q ]^{1/2}\frac{32 R^3 \nu(\hat{J}_\lambda,w_0)^4}{\lambda^{3/2}} \\
\label{eq:A1}  & \leqslant &
  \frac{ R   [ q(Q+\lambda \idm)^{-1} q ]^{1/2}}{ \lambda^{1/2} }
    \frac{32 R^2 \nu(\hat{J}_\lambda,w_0)^4}{\lambda} .
  \EEA
  Moreover, we have from the closed-form expression of $\hat{w}_\lambda^N$:
  \BEQ
  \label{eq:B1}
\big|  q^\top( \hat{w}_\lambda^N - w_0 ) - \frac{d_1}{n} \big| \leqslant
\big| \tr (Q+\lambda \idm)^{-1} ( qq^\top - Q/n) \big| + \lambda w_0^\top  (Q+\lambda \idm)^{-1} q.
  \EEQ
  Finally, we have, using \eq{boundgradientF} from Proposition~\ref{prop:multivariate}:
\BEA
\nonumber
\big|  q^\top \hat{J}_\lambda''(\hat{w}^N_\lambda)^{-1}
\hat{J}_\lambda'(\hat{w}^N_\lambda)   \big|
%& = &  \big|  q^\top \hat{J}_\lambda''(\hat{w}^N_\lambda)^{-1}
%[ \hat{J}_0'(\hat{w}^N_\lambda)  + \lambda  \hat{w}^N_\lambda
%-\hat{J}_0'(w_0) - \lambda w_0- (Q + \lambda \idm) (\hat{w}^N_\lambda-w_0)
%]\big| \\
& \!\!=\!\! &  \big|  q^\top \hat{J}_\lambda''(\hat{w}^N_\lambda)^{-1}
[ \hat{J}_0'(\hat{w}^N_\lambda)  
-\hat{J}_0'(w_0)  -  Q (\hat{w}^N_\lambda-w_0)
]\big| \\
\nonumber &\!\! \leqslant\!\! & \big[
 q^\top \hat{J}_\lambda''(\hat{w}^N_\lambda)^{-1} Q  \hat{J}_\lambda''(\hat{w}^N_\lambda)^{-1} q \big]^{1/2}  
  \big[
 \Delta^\top Q  \Delta\big]^{1/2}   R \| \Delta\|_2 \\
\label{eq:C1} & \!\!\leqslant\!\! & 2 \big[
 q^\top   Q   (Q+\lambda \idm)^{-2} q \big]^{1/2}  
  \| Q^{1/2}  \Delta \|_2  \frac{R \nu(\hat{J}_\lambda,w_0) }{\lambda^{1/2}},
  \EEA
  where $\Delta = \hat{w}^N_\lambda-w_0$. 
  
  What also needs to be shown is that 
  $ \big| \tr \hat{Q}_\lambda ( \hat{Q}_\lambda  + \lambda \idm)^{-1} - \tr Q (Q  + \lambda \idm)^{-1}  \big|$ is small enough; by noting that $Q=J_0''(w_0)$, $\hat{Q}_\lambda = J_0''(w_0 + v)$, and $v = \hat{w}_\lambda - w_0$, we have, using \eq{new} from Appendix~\ref{app:multivariate}:
  \BEA
  \nonumber
  & & \big| \tr \hat{Q}_\lambda ( \hat{Q}_\lambda  + \lambda \idm)^{-1} - \tr Q (Q  + \lambda \idm)^{-1}  \big| \\
\nonumber  & = & \lambda \big| \tr   \big[ ( \hat{Q}_\lambda  + \lambda \idm)^{-1}  ( Q - \hat{Q}_\lambda )  (Q  + \lambda \idm)^{-1} \big]  \big| \\
  \nonumber& \leqslant & \lambda \sum_{i=1}^p
  \big| \delta_i^\top
  ( \hat{Q}_\lambda  + \lambda \idm)^{-1}  ( Q - \hat{Q}_\lambda )  (Q  + \lambda \idm)^{-1} \delta_i
  \big| \\
 \nonumber & \leqslant & \lambda R   \sum_{i=1}^p
  \| Q^{1/2}
  ( Q  + \lambda \idm)^{-1} \delta_i \|_2
   \|  
  ( \hat{Q}_\lambda  + \lambda \idm)^{-1} \delta_i \|_2
   \| Q^{1/2} v \|_2
\\
  \label{eq:D1} & \leqslant & \lambda^{-1/2} R   \| Q^{1/2} v \|_2 \sum_{i=1}^p
  \delta_i^\top Q ( Q+ \lambda \idm)^{-1} \delta_i =   \lambda^{-1/2} R   \| Q^{1/2} v \|_2 d_1. \EEA
  All the terms in Eqs.~(\ref{eq:A1},\ref{eq:B1},\ref{eq:C1},\ref{eq:D1}) that need to be added to obtain the required upperbound are essentially the same than the ones proof of Theorem~\ref{theo:perf} in Appendix~\ref{app:perf} (with smaller constants). Thus the rest of the proof follows.

\section{Proof of Theorem~\ref{theo:consistency}}
\label{app:consistency}

We follow the same proof technique than for the Lasso~\cite{martin,Zhaoyu,zou}, i.e., we consider $\tilde{w}$ the minimizer of $\hat{J}_0(w) + \lambda s^\top w $ subject to $ {w}_{K^c}=0$ (which is unique because $Q_{KK}$ is invertible), and (1) show that $\tilde{w}_{K}$ has the correct (non zero) signs and (2) that
it is actually the unrestricted minimum of $\hat{J}_0(w) + \lambda \| w\|_1$ over $\rb^p$, i.e., using optimality conditions for nonsmooth convex optimization problems~\cite{borlew}, that $\| [ \hat{J}_0'(\tilde{w}) ]_{K^c} \|_\infty \leqslant \lambda$. All this will be shown by replacing $\tilde{w}$ by the proper one-step Newton iterate from $w_0$.

\paragraph{Correct signs on $K$.}
We directly use Proposition~\ref{prop:multivariate-bis} with the function
$w_K \mapsto \hat{J}_0(w_K,0) + \lambda s_K^\top w_K$---where $(w_K,0)$ denotes the $p$-dimensional vector obtained by completing $w_K$ by zeros---to obtain from \eq{nu2}:
$$( \tilde{w}_K - (w_0)_K)^\top Q_{KK}
(\tilde{w}_K - (w_0)_K) \leqslant 16 ( q_K - \lambda s_K)^\top Q_{KK} ^{-1} ( q_K - \lambda s_K) = 16 \nu^2, $$
 as soon as
$ \nu^2 = ( q_K - \lambda s_K)^\top   Q_{KK}^{-1} ( q_K - \lambda s_K) \leqslant \frac{\rho }{4R^2}$, and thus as soon as
$   q_K  Q_{KK}^{-1}   q_K  \leqslant \frac{\rho }{8R^2}$ and
$ \lambda^2 s_K^\top Q_{KK}^{-1}s_K \leqslant \frac{\rho }{8R^2}$. 
We thus  have:
$$
\| \tilde{w} - w_0 \|_\infty
  \leqslant   \| \tilde{w}_K - (w_0)_K \|_2
  \leqslant     \rho^{-1/2}
\| Q_{KK}^{1/2}
(\tilde{w}_K - (w_0)_K) \|_2 \leqslant 4  \rho^{-1/2}  \nu. $$
We therefore get the correct signs for the covariates indexed by $K$, as soon as $\| \tilde{w} - w_0 \|_\infty^2 \leqslant
 \min_{j \in K } |(w_0)_j|^2 = \mu^2$, i.e., as soon as
$$ \max \left \{ q_K  Q_{KK}^{-1}   q_K,  \lambda^2 s_K^\top Q_{KK}^{-1}s_K  \right\} \leqslant \min \left\{
\frac{\rho}{16} \mu^2,
\frac{\rho}{8R^2}
\right\}. $$
Note that $ s_K^\top Q_{KK}^{-1}s_K   \leqslant |K| \rho^{-1}$, thus it is implied by
the following constraint:
\BEQ
\lambda \leqslant \frac{\rho  }{ 4 |K|^{1/2} }
\min \left\{
 \mu,
R^{-1}
\right\},\EEQ
\BEQ
q_K  Q_{KK}^{-1}   q_K \leqslant \frac{\rho }{ 16 }
\min \left\{ \mu^2,
R^{-2}
\right\}.
\EEQ

\paragraph{Gradient condition on $K^c$.}

We denote by $\tilde{w}^N$ the one-step Newton iterate from $w_0$ for the minimization
of $\hat{J}_0(w) + \lambda s^\top w $ restricted to $w_{K^c}=0$,
equal to $\tilde{w}^N_K = (w_0)_K + Q_{KK}^{-1} ( q_K - \lambda s_K)$. From \eq{nu4}, we get:
$$( \tilde{w}_K - \tilde{w}_K^N)^\top Q_{KK}
(\tilde{w}_K -\tilde{w}_K^N) \leqslant \frac{16R^2}{\rho} \big[  ( q_K - \lambda s_K)^\top Q_{KK} ^{-1} ( q_K - \lambda s_K) \big]^2 = \frac{16R^2 \nu^4}{\rho}. $$

We thus have 
\BEAS
\| \tilde{w}- \tilde{w}^N\|_2 & \leqslant & \rho^{-1/2}  \frac{4R \nu^2}{\rho^{1/2}}
=  \frac{4R \nu^2}{\rho} \leqslant 1/R, \\
\| w_0 - \tilde{w}^N\|_2 & \leqslant & \rho^{-1/2}  \nu \leqslant 1/2R ,\\
\|\tilde{w} - w_0\|_2 & \leqslant  & \| \tilde{w}- \tilde{w}^N\|_2 + \| w_0 - \tilde{w}^N\|_2 \leqslant 3 \nu \rho^{-1/2} \leqslant 3R/2.
\EEAS
Note that up to here, all bounds $R$ may be replaced by the maximal $\ell_2$-norm of all data points, reduced to variables in $K$.

In order to check the gradient condition, we compute the gradient of $\hat{J}_0$ along the directions in $K^c$, to obtain for all $z \in \rb^p$, using \eq{boundgradientF} and
with any $v$ such that $ R \| v\|_2 \leqslant 3/2$ :
$$
\frac{
\big| z ^\top [  \hat{J}_0'(w_0+v) \! - \! \hat{T}_0'(w_0+v)    ] \big| }
{ (z ^\top Q z)^{1/2}  }
\leqslant
  (  v^\top Q v)^{1/2} \frac{e^{  R \| v\|_2} \!-\! 1\! -\! R\|v\|_2}{  R \| v\|_2 }
  \leqslant 2
  (  v^\top Q v)^{1/2} R \| v\|_2  ,
$$
where $ \hat{T}_0'(w) =  \hat{J}_0'(w_0) + \hat{J}_0''(w_0)(w-w_0)$ is the derivative of the Taylor expansion of $\hat{J}_0$ around $w_0$.
This implies, since $\diag(Q)\leqslant 1/4$, the following $\ell_\infty$-bound on the difference $\hat{J}_0$ and its Taylor expansion:
$$
\| [  \hat{J}_0'(w_0+v)  -\hat{T}_0'(w_0+v)  ]_{K^c} \|_\infty \leqslant 
    (  v^\top Q v)^{1/2} R \| v\|_2 .
$$
We now have,
\BEAS
\| \hat{J}_0'(\tilde{w})_{K^c} \|_\infty \!\!\!
& \!\!\leqslant\!\! & 
\|   \hat{T}_0'(\tilde{w}^N)_{K^c}\|_\infty \\
& & 
+
\| \hat{T}_0'(\tilde{w}^N)_{K^c} -  \hat{T}_0'(\tilde{w})_{K^c}\|_\infty
+
\|   \hat{T}_0'(\tilde{w})_{K^c}- \hat{J}_0'(\tilde{w})_{K^c} \|_\infty,
\\
&\!\! \leqslant \!\!& 
 \| [ \hat{J}_0'(w_0) 
+  Q(\tilde{w}^N-w_0)   ]_{K^c}  \|_\infty \\
& & 
+ \| [ Q( \tilde{w} - \tilde{w}^N)   ]_{K^c}  \|_\infty
+  R \| \tilde{w}- w_0\|_2 \| Q^{1/2}(\tilde{w}- w_0)\|_2,
 \\
&\!\! \leqslant\!\! &  \|  -q_{K^c} + Q_{K^c K } Q_{KK}^{-1} ( q_K - \lambda s_K ) \|_\infty \\
& & 
+ \| Q_{K^c K} Q_{KK}^{-1/2} Q_{KK}^{1/2}( \tilde{w}_K - \tilde{w}^N_K)  \|_\infty
+ 
3 \nu R \rho^{-1/2} ( 4 R \nu^2 \rho^{-1/2} + \nu )
,
\\
&\!\! \leqslant \!\!&  \|  q_{K^c} - Q_{K^c K } Q_{KK}^{-1} ( q_K - \lambda s_K ) \|_\infty
+ \frac{1}{4} \| Q_{KK}^{1/2}( \tilde{w}_K - \tilde{w}^N_K)  \|_2
+   \frac{ 9  R}{\rho^{1/2} } \nu^2
,
\\
&\!\! \leqslant \!\!&  \|  q_{K^c} - Q_{K^c K } Q_{KK}^{-1} ( q_K - \lambda s_K ) \|_\infty
+ \frac{1}{4} \frac{16 R}{\rho^{1/2} } \nu^2 +   \frac{ 9  R}{\rho^{1/2} } \nu^2,
\\
& \!\!\leqslant \!\!&  \|  q_{K^c} - Q_{K^c K } Q_{KK}^{-1} ( q_K - \lambda s_K ) \|_\infty
+ \frac{16 R}{\rho^{1/2} } \nu^2.
 \EEAS
 Thus, in order to get $\| \hat{J}_0'(\tilde{w})_{K^c} \|_\infty \leqslant \lambda$,
 we need
\BEQ
\| q_{K^c} - Q_{K^c K } Q_{KK}^{-1}  q_K \|_\infty \leqslant   \eta \lambda/4,
\EEQ
and
\BEQ
\max \left \{ q_K  Q_{KK}^{-1}   q_K,  \lambda^2 s_K^\top Q_{KK}^{-1}s_K  \right\} \leqslant \frac{ \lambda \eta \rho^{1/2} }{64 R }.
\EEQ
In terms of upper bound on $\lambda$ we then get:
$$
\lambda \leqslant 
\min \left\{
 \frac{\rho  }{ 4 |K|^{1/2} }\mu,
\frac{\rho  }{ 4 |K|^{1/2} }R^{-1}
,
 \frac{ \eta \rho^{3/2} }{64 R |K|}
 \right\},\
$$
which can be reduced
$
\lambda \leqslant 
\min \left\{
 \frac{\rho  }{ 4 |K|^{1/2} }\mu,
 \frac{ \eta \rho^{3/2} }{64 R |K|}
 \right\}
$.
In terms of upper bound on $q_K^\top Q_{KK}^{-1} q_K$ we get:
$$
q_K^\top Q_{KK}^{-1} q_K  \leqslant \min \left\{ \frac{\rho }{ 16 }
 \mu^2,
\frac{\rho }{ 16 }
R^{-2},  \frac{\lambda \eta \rho^{1/2} }{64 R }
\right\},
$$
which can be reduced to
$
q_K^\top Q_{KK}^{-1} q_K  \leqslant \min \left\{ \frac{\rho }{ 16 }
 \mu^2,
 \frac{\lambda \eta \rho^{1/2} }{64 R }
\right\}$, using the constraint on $\lambda$.

We now derive and use concentration inequalities. We first use Bernstein's inequality (using for all $k$ and $i$,
$| (x_i)_k - Q_{k K } Q_{KK}^{-1}  (x_i)_K | | \varepsilon_i| \leqslant R/\rho^{1/2}$
and $Q_{kk} \leqslant 1/4$),
 and the union bound to get
\BEAS
\P( \| q_{K^c} - Q_{K^c K } Q_{KK}^{-1}  q_K \|_\infty \geqslant  \lambda \eta / 4)
& \!\leqslant \! &  2 p\exp \left(
- \frac{ n \lambda^2 \eta^2 / 32 }{ 1/4 +  R\lambda \eta \rho^{-1/2} / 12}
\right)  \\
& \!\leqslant \! & 2 p\exp \left(
- \frac{n \lambda^2 \eta^2 }{16}
\right),
\EEAS
as soon as $    R\lambda \eta \rho^{-1/2}   \leqslant 3$,
i.e., as soon as, $\lambda \leqslant  3 \rho^{1/2} R^{-1}$, which is indeed satisfied because of our assumption on $\lambda$.
We  also use Bernstein's inequality to get
$$
\P( q_K^\top Q_{KK}^{-1} q_K \geqslant t)
\leqslant \P \bigg(  \| q_K \|_\infty \geqslant \sqrt{\frac{\rho t}{|K|}}  \bigg)
\leqslant 2 |K| \exp\bigg( - \frac{n\rho t}{|K|}  \bigg).
$$ 
The union bound then leads to the desired result.

\section{Proof of Theorem~\ref{theo:efficiency}}
\label{app:efficiency}
We follow the proof technique of \cite{tsyb}.
We have $
\hat{J}_0(\hat{w}_\lambda) = J_0(\hat{w}_\lambda) - q^\top  \hat{w}_\lambda $. Thus, because $\hat{w}_\lambda$ is a minimizer of $\hat{J}_0(w) + \lambda \| w\|_1$,
\BEQ
\label{eq:a}
 {J}_0(\hat{w}_\lambda)   - q^\top \hat{w}_\lambda  
 + \lambda \|\hat{w}_\lambda \|_1 \leqslant {J}_0(w_0)  -q^\top w_0
 + \lambda \|w_0 \|_1,
\EEQ
which implies, since ${J}_0(\hat{w}_\lambda) \geqslant J_0(w_0)$:
\BEAS
 \lambda \|\hat{w}_\lambda \|_1 & \!\!\! \leqslant\! \!\! &\lambda \|w_0 \|_1 
+ \| q\|_\infty \|\hat{w}_\lambda - w_0\|_1,
\\
  \lambda \|(\hat{w}_\lambda)_K \|_1
 + \lambda \|(\hat{w}_\lambda)_{K^c} \|_1 &  \!\!\! \leqslant \!\!\!   & \lambda \|(w_0)_{K} \|_1 
+ \| q\|_\infty \big(
\| (\hat{w}_\lambda)_K - (w_0)_K \|_1
+\| (\hat{w}_\lambda)_{K^c}\|_1
\big).
 \EEAS
 If we denote by $\Delta =  \hat{w}_\lambda - w_0 $ the estimation error, we deduce:
 $$ ( \lambda - \| q\|_\infty ) \| \Delta_{K^c} \|_1
 \leqslant  ( \lambda + \| q\|_\infty )  \| \Delta_{K} \|_1.
 $$
 If we assume $\| q\|_\infty \leqslant \lambda/2$, then, we have
 $ \| \Delta_{K^c} \|_1 \leqslant 3 \| \Delta_{K} \|_1$, and thus using \textbf{(A5)},
 we get $\Delta^\top Q \Delta \geqslant \rho^2 \| \Delta_K \|_2^2$.
 From \eq{a}, we thus get:
 \BEA
 \nonumber
 {J}_0(\hat{w}_\lambda) -  {J}_0(w_0) &   \leqslant &   q^\top(\hat{w}_\lambda - w_0)
 - \lambda \|\hat{w}_\lambda \|_1 
 + \lambda \|w_0 \|_1,
\\
\label{eq:AAAAAA}
 {J}_0(w_0 + \Delta ) -  {J}_0(w_0) & \leqslant &  ( \| q\|_\infty + \lambda ) \| \Delta\|_1
 \leqslant \frac{3 \lambda}{2} \| \Delta\|_1.
 \EEA
 Using \eq{lowerboundF} in Proposition~\ref{prop:multivariate} with $J_0$, we obtain:
$$ {J}_0(w_0 + \Delta ) -  {J}_0(w_0)  \geqslant \frac{ \Delta^\top Q \Delta}{R^2 \| \Delta\|_2^2}
\big(
e^{-R \|\Delta\|_2} + R \|\Delta\|_2 - 1
\big), 
 $$
 which implies, using $\Delta^\top Q \Delta \geqslant \rho^2 \| \Delta_K\|_2^2$ and \eq{AAAAAA}:
 \BEQ
 \label{eq:eq}
  \frac{ \rho^2 \| \Delta_K \|_2^2}{R^2 \| \Delta\|_2^2}
\big(
e^{-R \|\Delta\|_2} + R \|\Delta\|_2 - 1
\big)
 \leqslant \frac{3 \lambda}{2} \| \Delta\|_1 .
 \EEQ
 We can now use, with $s = |K|$, $\| \Delta\|_2 \leqslant \|\Delta\|_1 \leqslant 4 \|\Delta_K\|_1 \leqslant 4 \sqrt{s} \| \Delta_K\|_2 $ 
 to get:
  $$
   { \rho^2  } 
\big(
e^{-R \|\Delta\|_2} + R \|\Delta\|_2 - 1
\big)
\leqslant \frac{3\lambda}{2} \frac{ ( 4 \sqrt{s} \| \Delta_K\|_2 )^2 R \|\Delta\|_2}{\|\Delta_K\|_2^2}
 \leqslant  {24 \lambda{s}} R^2 \|\Delta\|_2.
 $$
 This implies using \eq{kappa}, that $R\|\Delta\|_2 \leqslant \frac{ 48 \lambda R s /\rho^2}{ 1- 24 \lambda s R /\rho^2 } \leqslant 2$ a soon as $R\lambda s \rho^{-2} \leqslant 1 / 48$, which itself implies that 
 $
 \frac{1}{(R  \| \Delta\|_2)^2}
\big(
e^{-R \|\Delta\|_2} + R \|\Delta\|_2 - 1
\big) \geqslant 1/2
$, and thus, from \eq{eq},  
$$ \| \Delta_K\|_2 \leqslant \frac{3 \lambda}{2} \times
4 \sqrt{s} \| \Delta_K\|_2  .$$
The second result then follows from \eq{AAAAAA} (using Bernstein inequality for an upper bound on $\P(\|q\|_\infty \geqslant \lambda/2)$).

 \section{Concentration inequalities}
 \label{sec:concentration}
 \label{app:concentration}
  In this section, we derive concentration inequalities for quadratic forms of bounded random variables that extend the ones already known for Gaussian random variables~\cite{minikernel}. The following proposition is a simple corollary of a general concentration result on U-statistics~\cite{reynaud}.

\begin{proposition}
\label{prop:concentration}
Let $y_1,\dots,y_n$ be $n$  vectors in $\rb^p$ such that $\| y_i\|_2 \leqslant b$ for all $i=1,\dots,n$ and $Y = [ y_1^\top, \dots, y_n^\top ]^\top \in \rb^{n\times p }$. Let $\varepsilon \in \rb^n $ be a vector of zero-mean independent random variables almost surely bounded by 1 and with variances $\sigma^2_i$, $i=1,\dots,n$. Let $S = \Diag(\sigma_i)^\top Y Y^\top \Diag(\sigma_i)$. Then, for all $u \geqslant 0$:
\begin{multline}
\P
\big[ \
 | \varepsilon ^\top YY^\top \varepsilon -  \tr S | \geqslant
32 \tr (S^2)^{1/2} u^{1/2} 
+ 18\lmax(S) u  \\
+ 126 b (\tr S)^{1/2} u^{3/2} 
+ 39 b^2 u^2
\big] \leqslant 8 e^{-u}.
\end{multline}
\end{proposition}
\begin{proof}
We apply Theorem 3.4 from~\cite{reynaud}, with $T_i = \varepsilon_i$,
$g_{i,j}(t_i,t_j) = y_i^\top y_j  t_i t_j$ if $|t_i|, |t_j| \leqslant 1$ and zero otherwise. We then have (following notations from~\cite{reynaud}):
\BEAS
A & = & \max_{i,j} | y_i^\top y_j | \leqslant b^2, \\
B^2 & = & \max_{i \in \{1,\dots,n \}}
\sum_{j<i} ( y_i^\top y_j )^2 \sigma_j^2
\leqslant \max_{i \in \{1,\dots,n \}}
\sum_{j<i}  y_i^\top y_i b^2 \sigma_j^2
\leqslant b^2 \tr(S), \\
C^2 & =  & \sum_{j < i }  ( y_i^\top y_j )^2 \sigma_j^2 \sigma_i^2
\leqslant  \frac{1}{2} \tr (S^2), \\
D & \leqslant & \frac{1}{2} \lmax( S ) .
\EEAS
Thus (using $\varepsilon = 4$ in \cite{reynaud}):
$$
\P \bigg(  \bigg| \sum_{j\neq i} y_i^\top y_j \varepsilon_i 
 \varepsilon_j \bigg| \geqslant 44.8 C u^{1/2} + 35.36 Du + 124.56 B u^{3/2} + A 38.26 u^2 \bigg) \leqslant 5.54 2 e^{-u}.
$$
Moreover, we have from Bernstein's inequality~\cite{massart-concentration}:
$$\P \bigg(  \bigg| \sum_{i=1}^n y_i^\top y_i ( \varepsilon_i^2 - \sigma_i^2) \bigg|
\geqslant  u^{1/2} \sqrt{ 2 b^2 \tr S  } + \frac{b^2 u }{3}   \bigg) \leqslant 2  e^{-u },
$$
leading to the desired result, noting that for $u \leqslant \log(8)$, the bound is trivial.
\end{proof}

 We can apply to our setting to get, with 
 $y_i = \frac{1}{n} ( P+ \lambda \idm)^{-1/2} x_i$ (with $\|x_i\|_2 \leqslant R$), leading to $b = \frac{1}{2} R n^{-1} \lambda^{-1/2}$ and $S = \frac{1}{n}\Diag(\sigma) X ( P+ \lambda \idm)^{-1} X^\top \Diag(\sigma)$.

 \paragraph{Misspecified models.} If no assumptions are made, we  simply have:
 $ \lmax(S) \leqslant ( \tr S^2)^{1/2} \leqslant \tr(S) \leqslant R^2/ \lambda n$ and we get  after bringing terms together:
   \BEQ
\label{eq:concentration-no-assumptions}
 \P \bigg[  q^\top( P + \lambda\idm)^{-1} q   \geqslant
 \frac{41 R^2 u }{\lambda n} 
 +
 \frac{R^2}{\lambda} \bigg( 8 \frac{u^2}{n^2} + 63\frac{u^{3/2}}{n^{3/2}} \bigg)
 \bigg]
  \leqslant 8 e^{-u}.
 \EEQ

\paragraph{Well-specified models} In this case, $P=Q$ and $\lmax(S)  \leqslant 1/n$,
$\tr S = d_1 /n$, $\tr S^2 = d_2 / n^2$.
  \BEQ
\label{eq:concentration-wellspecified}
 \P \bigg[   \bigg| q^\top( P + \lambda\idm)^{-1} q - \frac{d_1}{n}    \bigg| \geqslant \frac{32 d_2^{1/2} u^{1/2}}{n } + \frac{18u}{n} + \frac{ 53 R d_1^{1/2} u^{3/2}}{n^{3/2}\lambda^{1/2}} + 9 \frac{R^2 u^2}{\lambda n^2}
 \bigg]
  \leqslant 8 e^{-u}.
 \EEQ

\section*{Acknowledgements}

I would like to thank Sylvain Arlot,  Jean-Yves Audibert and Guillaume Obozinski   for fruitful discussions related to this work. This work was supported by a French grant from the Agence Nationale de la Recherche (MGA Project ANR-07-BLAN-0311).

\bibliography{selfconcordant}
\end{document}